%% file: main.tex
\documentclass[lettersize,journal]{IEEEtran}
\usepackage{amsmath,amsfonts}
\usepackage{array}
\usepackage[caption=false,font=normalsize,labelfont=sf,textfont=sf]{subfig}
\usepackage{textcomp}
\usepackage{stfloats}
\usepackage{url}
\usepackage{verbatim}
\usepackage{graphicx}
\usepackage{cite}
\usepackage[numbers,longnamesfirst]{natbib}
\usepackage{smile}
\usepackage{xcolor}
\usepackage{physics}
\newcommand{\diff}{\mathrm{d}}
\usepackage{mathtools}
\usepackage[ruled,vlined]{algorithm2e}
\title{Self-Guidance: Boosting Flow and Diffusion Generation on Their Own}

\author{Tiancheng Li$^{1,2}$\quad Weijian Luo$^{2,3,*}$  \quad Zhiyang Chen$^{2,4}$\quad Liyuan Ma$^{2,4}$\quad Guo-Jun Qi$^{2,4,*}$\\
    $^1$Zhejiang University, $^2$MAPLE Lab, Westlake University, $^3$Peking University\\
    $^4$Institute of Advanced Technology, Westlake Institute for Advanced Study\\
    \tt\small \{litiancheng, chenzhiyang, maliyuan\}@westlake.edu.cn\\
\tt\small luoweijian@stu.pku.edu.cn, guojunq@gmail.com
}
\begin{document}
\maketitle
\let\thefootnote\relax\footnotetext{This project was initiated and supported by MAPLE Lab at Westlake University.}
\let\thefootnote\relax\footnotetext{$^*$ Corresponding author.}


\input{sec/0_abstract}    
\input{sec/1_intro}

\input{sec/2_relatedwork}

\input{sec/3_preliminary}

\input{sec/4_connection}

\input{sec/5_experiment}

\input{sec/6_conclusion}

{
    \footnotesize
    \bibliographystyle{ieeenat_fullname}
    \bibliography{main}
}

\clearpage
\input{sec/X_suppl}

\end{document}

%% file: sec/0_abstract.tex
\begin{abstract}
Proper guidance strategies are essential to achieve high-quality generation results without retraining diffusion and flow-based text-to-image models. Existing guidance either requires specific training or strong inductive biases of diffusion model networks, which potentially limits their ability and application scope. 
Motivated by the observation that artifact outliers can be detected by a significant decline in the density from a noisier to a cleaner noise level, we propose Self-Guidance (SG), which can significantly improve the quality of the generated image by suppressing the generation of low-quality samples. The biggest difference from existing guidance is that SG only relies on the sampling score function of the original diffusion or flow model at different noise levels, with no need for any tricky and expensive guidance-specific training. This makes SG highly flexible to be used in a plug-and-play manner by any diffusion or flow models. 
We also introduce an efficient variant of SG, named SG-prev, which reuses the output from the immediately previous diffusion step to avoid additional forward passes of the diffusion network.
We conduct extensive experiments on text-to-image and text-to-video generation with different architectures, including UNet and transformer models. 
With open-sourced diffusion models such as Stable Diffusion 3.5 and FLUX, SG exceeds existing algorithms on multiple metrics, including both FID and Human Preference Score. SG-prev also achieves strong results over both the baseline and the SG, with 50 percent more efficiency. 
Moreover, we find that SG and SG-prev both have a surprisingly positive effect on the generation of physiologically correct human body structures such as hands, faces, and arms, showing their ability to eliminate human body artifacts with minimal efforts. 
We have released our code at \url{https://github.com/maple-research-lab/Self-Guidance}.

\end{abstract}


%% file: sec/1_intro.tex
\section{Introduction}\label{sec:intro}
\begin{figure}
    \centering
    \includegraphics[width=1\linewidth]{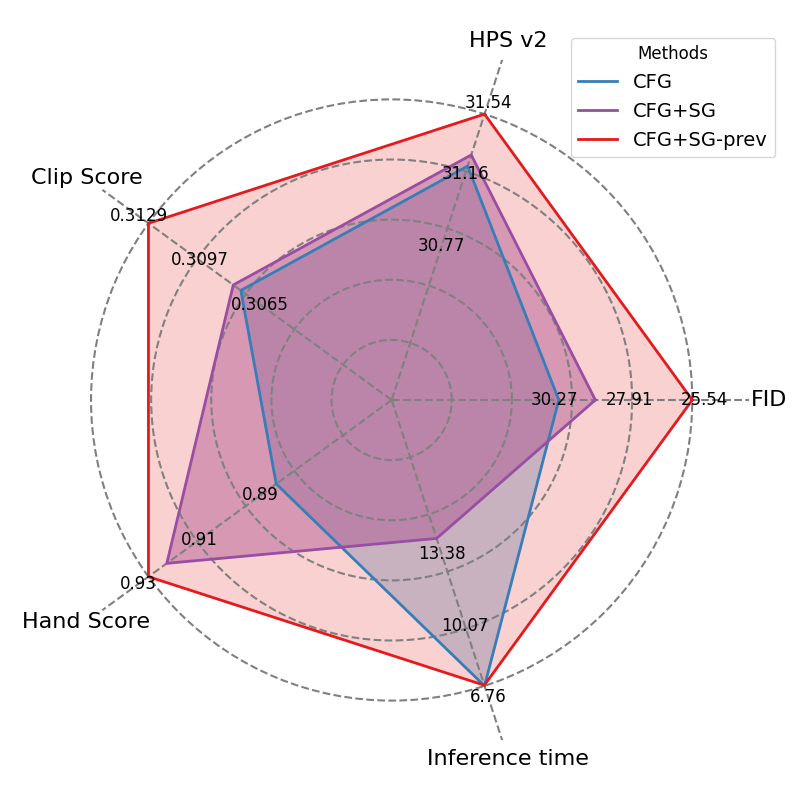}
    \vspace{-5mm}
    \caption{Radar chart of 5 image benchmarks on Flux.1. Background numbers indicate reference values corresponding to normalized radii (0.5, 0.75, 1.0)}
    \label{fig:radar}
    \vspace{-5mm}
\end{figure}
\begin{figure*}
    \centering
    \includegraphics[width=1\linewidth]{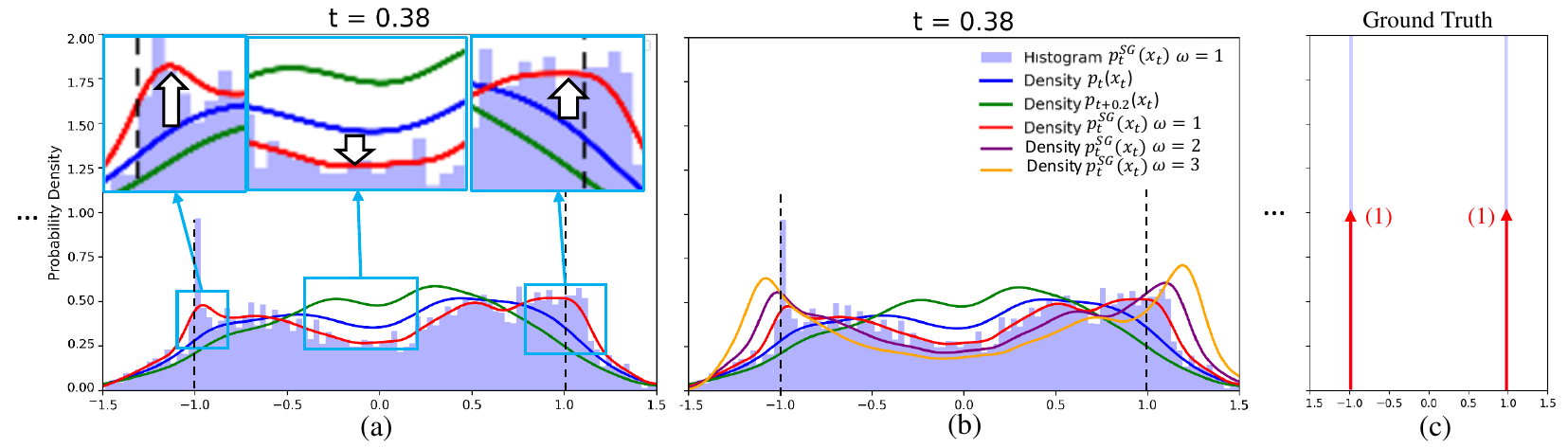}
    \vspace{-5mm}
    \caption{We train a diffusion model on a 1-dimensional toy example with data drawn from two separate modes at $\pm 1$. At different noise levels $t$, we fit and plot the distribution of generated samples in the reverse diffusion process. (a) The results on diffusion time $t=0.38$. The blue curve plots the distribution of generated samples at the current noise level $t$, while the green curve plots the distribution of samples generated at the noisier level $t+\delta(t)$. The red curve shows the obtained SG sampling distribution after applying the ratio of these two probabilities with $\omega=1$. (b) With various $\omega$, we show that artifact samples around the origin are suppressed more sharply with a larger value of $\omega$, while the density of samples around two groundtruth modes is boosted. (c) The ground truth two-mode distribution at $\pm 1$. 
    }
    \label{fig:two_point_distribution}
    \vspace{-5mm}
\end{figure*}

\IEEEPARstart{O}{ver} the past decade, deep generative models have achieved remarkable advancements across various applications \citep{karras2020analyzing,karras2022edm,nichol2021improved,oord2016wavenet,ho2022video,poole2022dreamfusion,hoogeboom2022equivariant,kim2021guidedtts,kingma2018glowgenerativeflowinvertible,chen2019residual,meng2021sdedit, Couairon2022DiffEditDS,luo2024diffstar,luo2024diffinstruct,luo2024diffpp,geng2024consistency,luo2024one}. Among them, diffusion models \citep{NEURIPS2020_4c5bcfec,song2020denoising,NEURIPS2021_0a9fdbb1} and flow-based models \citep{esser2024scalingrectifiedflowtransformers,stability2023stable_diffusion,blackforestlabs} have notably excelled in producing high-resolution, text-driven data such as images \citep{saharia2022photorealistic,ramesh2022hierarchical,ramesh2021zero}, videos \citep{ho2022video,blattmann2023stablevideodiffusionscaling,yang2024cogvideoxtexttovideodiffusionmodels}, and others \citep{kong2021diffwave,oord2016wavenet,shen2023difftalkcraftingdiffusionmodels,huang2024flow,luo2024entropy,zhang2023enhancing,wang2024integrating,deng2024variational}, pushing the boundaries of Artificial Intelligence Generated Contents.

In simple terms, diffusion models learn a multi-step transition from a prior distribution $p_T(\bx_T)$ to a real data distribution $p_0(\bx_0)$.
However, default sampling methods for diffusion and flow-based models often lead to unsatisfactory generation quality, such as broken human hands and faces, and images with bad foreground and background. To address these issues, various guidance strategies have emerged as cheap yet effective ways to guide the generation process for better generation quality. For instance, classifier-free guidance (CFG) \cite{ho2022classifierfreediffusionguidance} modifies the velocity of diffusion and flow-based generative models by adding a delta term between class-conditional and unconditional velocities, which pushes generated samples to have high class probabilities.

Though these existing guidance have shown impressive performance improvements, they have various individual restrictions. For instance, the CFG relies on computing an additional unconditional velocity, which requires training the diffusion model under both conditional and unconditional settings, therefore, harms the modeling performances\citep{ahn2024selfrectifyingdiffusionsamplingperturbedattention}. Auto-Guidance (AG) pays a significant price that requires training an additional \emph{bad-version} model, which is tricky as well as requiring more memory costs. Other guidance, such as Perturbed-attention Guidance (PAG \citep{ahn2024selfrectifyingdiffusionsamplingperturbedattention}), and self-attention guidance (SAG \citep{hong2023improvingsamplequalitydiffusion}), do not rely on additional training. However, as the PAG paper described, the effectiveness of PAG is highly sensitive to the selection of perturbed attention layers inside the neural network, making it less flexible to enhance models in general applications. We notice that \emph{all these guidance variants focus on diffusion sampling strategies at a single timestep, while neglecting how diffusion sampling at various timesteps could be explored to improve the generation quality}. Furthermore, we find that without explicitly retraining or perturbing an existing diffusion model as in CFG and PAG, using the diffusion samplings from different timesteps yields a Self-Guidance (SG) approach. 

Specifically, for a diffusion model, we find that the reason for generating unnatural images is due to the inadequate denoising that could yield artifacts in the reverse diffusion process.
As Figure \ref{fig:two_point_distribution} shows, let us consider a simple diffusion model trained on a 1-dimensional toy dataset with two separate modes at $\pm 1$ as the ground truth.
At the noise level $t=0.38$ for example, the probability  $p_t(x_t)$ (i.e., the blue curve) is likely to draw artifact samples around the origin as it has a large density at $x_t=0$.  

To relieve it, an intuitive idea is to take a look at the diffused probability $p_{t+\delta(t)}(x_t)$ at a noisier level $t+\delta(t)$ (i.e., the green curve). We know the samples at this level could be noisier and thus contain more severe artifacts. Therefore, by comparing between these two levels, one can assume that if the density at the cleaner level $t$ (although it is still high) declines significantly from that at the noisier level $t+\delta(t)$, the corresponding samples are likely to be artifacts, since the belief of sampling them as cleaner outputs has greatly dropped as reflected by the declining density.  This is what we have seen in Figure \ref{fig:two_point_distribution}(a)-(b), where the current sampling probability $p_t(x_t)$ of the blue curve at $t=0.38$ declines from that of the green curve at the noisier level $t+0.2$ at the origin.  This reveals that the samples around the origin could be artifacts, which is true. This decline can be measured by the ratio of these two probabilities, resulting in a new sampling strategy 
\begin{equation}
p^{SG}_t(x_t)\propto p_t(x_t)\big\{\frac{p_t(x_t)}{p_{t+\delta(t)}(x_t)}\big\}^{\omega},
\end{equation}
to guide the reverse diffusion process. Here, $\omega$ is the guidance scale. The higher the value of $\omega$, the more sharply the artifact samples will be suppressed as shown in Figure \ref{fig:two_point_distribution}(b).

\begin{figure*}[ht!]
    \centering
    \includegraphics[width=0.95\linewidth]{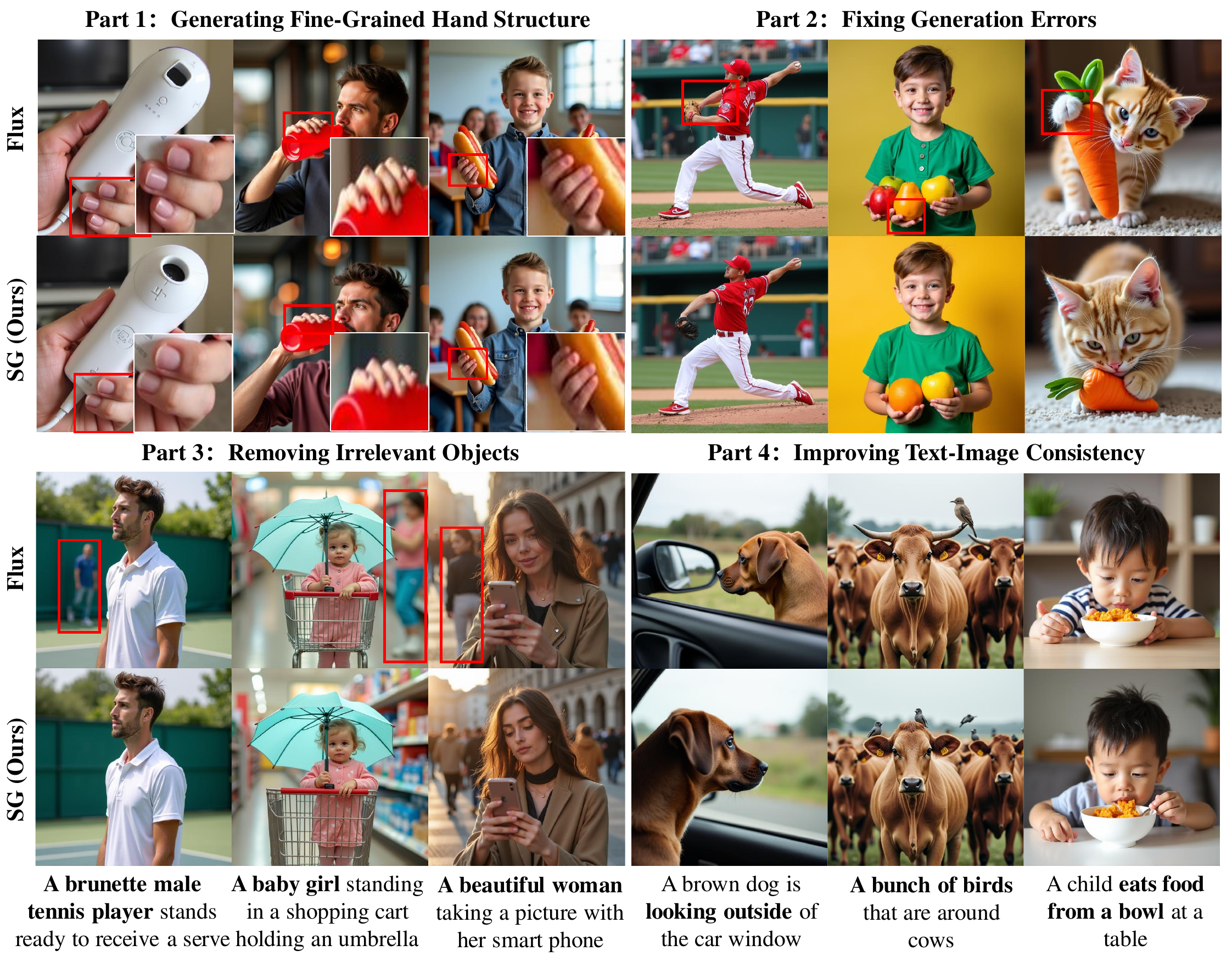}
    \vspace{-5mm}
    \caption{Qualitative comparisons between Flux.1 (baseline) and Self-Guidance (SG) diffusion samples. We compare the generated images of Flux and our SG from four parts. The red box in the figure represents the bad Flux generation and the better SG generation, and the enlarged image is shown in the lower right corner.}

    \label{fig:specific_task_generation}
    \vspace{-5mm}
\end{figure*}

In this paper, we present such an inference-time sampling strategy called \emph{Self-Guidance (SG)} since it only involves its own diffusion model at different noise levels.
For image generation tasks, as shown in Fig.~\ref{fig:specific_task_generation}, compared with the benchmark diffusion sampling algorithm, applying SG successfully removes unwanted artifacts in generated images,  fixing errors on human fingers and other generation errors, eliminating irrelevant objects, and improving text-image consistency.

This guidance can further be approximated using the output from the immediately previous diffusion step -- i.e., $p_{t+1}(x_{t+1})$. We denote this approximation as \emph{SG-prev}. SG-prev only needs one forward pass, thereby incurring no additional inference cost and keeping the overall inference time almost unchanged.

In addition, SG is architecture-independent and can therefore be integrated into almost all existing diffusion and flow-based models regardless of varying model architectures. Unlike CFG \cite{ho2022classifierfreediffusionguidance} and AG \cite{karras2024guidingdiffusionmodelbad}, SG does not require additional guidance-specific training. SG is orthogonal to many other guidance approaches \citep{ho2022classifierfreediffusionguidance,karras2024guidingdiffusionmodelbad,ahn2024selfrectifyingdiffusionsamplingperturbedattention}, seamlessly works together with them in a plug-and-play manner for better performances. 

In Section \ref{sec:5.2}, we apply SG and SG-prev on Flux.1-dev \cite{blackforestlabs} and Stable-Diffusion 3.5 \citep{rombach2022high,esser2024scalingrectifiedflowtransformers}, as well as a representative text-to-video model, CogVideoX\cite{yang2024cogvideoxtexttovideodiffusionmodels}. We show that when using SG only,  all models achieved improved FID \cite{heusel2018ganstrainedtimescaleupdate} and higher human preference scores on the HPS v2.1\cite{wu2023humanpreferencescorev2} benchmark. Combining SG and SG-prev with other established guidance methods, including CFG and PAG, leading to new state-of-the-art diffusion generation results, achieving 20.74 on FID, 31.56 on HPSv2, and 5.7845 on Aesthetic Score.
Experiments on text-to-video generation also confirm the solid performance improvements brought by SG.

The main contributions of this work are as follows.  
\begin{itemize}
    \item We observe that the ratio of probabilities at different noise levels serves as a good indicator of artifacts during the reverse diffusion process.
    \item Leveraging this insight, we propose \emph{Self-Guidance}—a plug-and-play guidance method that improves generation quality without requiring extra training, supervision, or external models. Moreover, such guidance can be implemented without incurring additional inference cost, using a variant we refer to as SG-prev.
    \item We show that SG and SG-prev are compatible with other sampling methods like CFG and PAG, leading to state-of-the-art image generation performances. 
\end{itemize}

%% file: sec/2_relatedwork.tex
\section{Related Works}

There are roughly two kinds of diffusion guidance. The first kind mixes the outputs of multiple models to get the generative velocity. The classifier-free guidance (CFG) \citep{ho2022classifierfreediffusionguidance} uses a mix of conditional and unconditional score functions as the velocity. Most recently, the Auto-guidance (AG) \citep{karras2024guidingdiffusionmodelbad} proposed to mix the current score output with a \emph{bad-version} model to enhance persistence of velocity, resulting in improved performances. Both CFG and AG use two models for inference explicitly or implicitly. Some other studies have also elaborately studied guidance through the lens of diffusion solvers \citep{lu2022dpm,liu2022pseudo,xue2024sa}.

The second line of guidance defines the velocity by mixing the output of the score functions of a perturbed and the original sample. These methods usually need one model for inference. The self-attention guidance (SAG) \citep{hong2023improvingsamplequalitydiffusion} proposed to assign Gaussian perturbations to samples for guidance. Recently, Perturbed-attention guidance (PAG) \citep{ahn2024selfrectifyingdiffusionsamplingperturbedattention} found that modifying the self-attention map of the diffusion model network for guidance results in strong performances. PAG-like guidance has shown steady improvements. Another recent work \citep{sadat2024trainingproblemrethinkingclassifierfree} proposed the TimeStep Guidance (TSG), which perturbed the timestep embedding of diffusion models, sharing a similar spirit as PAG. For perturbation-based guidance, how to choose proper perturbations is mostly decided in an ad-hoc manner, which may vary significantly across different neural network architectures. Other works also study perturbation-based guidance paradigms from different perspectives \citep{alemohammad2024selfimprovingdiffusionmodelssynthetic,hong2024smoothedenergyguidanceguiding,10.1145/3664647.3681506}. 

Despite improvements in overall image structure and class-label alignment, existing guidelines have a common problem: they still generate images with artifacts, like six-finger hands or twisted human bodies, which brings significant concerns in AI safety and user experiences. 
As a comparison, our proposed Self-Guidance significantly improves fine-grained details of generated images, completing weaknesses of existing guidance in an orthogonal way. We will introduce details of SG in Section \ref{sec:method}.

%% file: sec/3_preliminary.tex
\section{Preliminary}\label{sec:pre}
\begin{figure}
    \centering
    \includegraphics[width=0.8\linewidth]{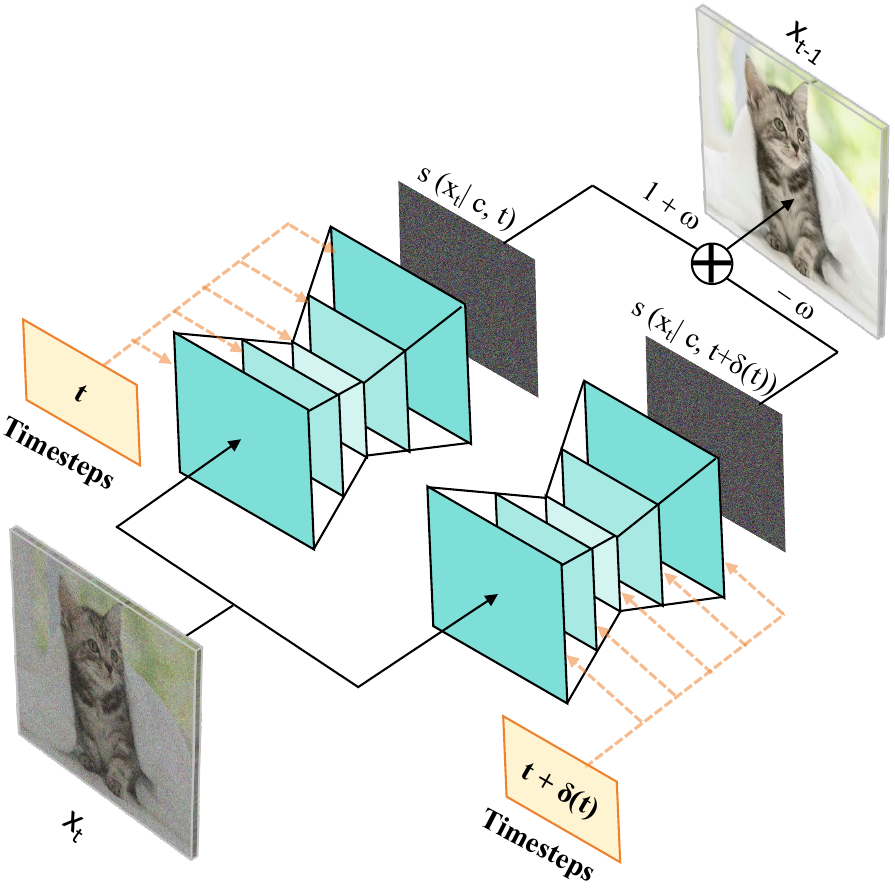}
    \caption{An illustration of Self-Guidance of one iteration step during generation.}
    \label{fig:structure}
    \vspace{-5mm}
\end{figure}
\paragraph{Diffusion Models} 
In this section, we introduce preliminary knowledge and notations about diffusion models \citep{song2021scorebased,ho2020denoising,Dickstein11}. The flow models \citep{liu2022flow,Lipman2022FlowMF} share similar concepts that we put in Appendix \ref{app:b}. Assume we observe data from the underlying distribution $q_d(\bx)$. 
The goal of generative modeling is to train models to generate new samples $\bx\sim q_d(\bx)$. 
The forward diffusion process of DM transforms $q_{0}=q_d$ towards some simple noise distribution, 
\begin{align}\label{equ:forwardSDE}
    \diff \bx_t = \bm{F}(\bx_t,t)\mathrm{d}t + G(t)\diff \bm{w}_t,
\end{align}
where $\bm{F}$ is a pre-defined drift function, $G(t)$ is a pre-defined scalar-value diffusion coefficient, and $\bm{w}_t$ denotes an independent Wiener process. 
A continuous-indexed score network $\bm{s}_\varphi (\bx,t)$ is employed to approximate marginal score functions of the forward diffusion process \eqref{equ:forwardSDE}. The learning of score networks is achieved by minimizing a weighted denoising score matching objective \citep{vincent2011connection, song2021scorebased},
\begin{align}\label{def:wdsm}
    \mathcal{L}_{DSM}(\varphi) = & \int_{t=0}^T \lambda(t) \mathbb{E}_{\bx_0 \sim q_{0}, \bx_t | \bx_0 \sim q_t(\bx_t | \bx_0)} \nonumber \\
    & \|\bm{s}_\varphi(\bx_t, t) - \nabla_{\bx_t} \log q_t(\bx_t | \bx_0)\|_2^2 \mathrm{d}t.
\end{align}

Here, the weighting function $\lambda(t)$ controls the importance of the learning at different time levels and $q_t(\bx_t|\bx_0)$ denotes the conditional transition of the forward diffusion \eqref{equ:forwardSDE}. 
After training, the score network $\bm{s}_{\varphi}(\bx_t, t) \approx \nabla_{\bx_t} \log q_t(\bx_t)$ is a good approximation of the marginal score function of the diffused data distribution. High-quality samples from a DM can be drawn by simulating generative SDE \eqref{def:rev_sde} which is implemented by replacing the score function $\nabla_{\bx_t} \log q_t(\bx_t)$ with the learned score network \citep{song2021scorebased}. 
\begin{align}\label{def:rev_sde}
    \diff\bx_t = & \bigg\{\bm{F}(\bx_t, t) - \frac{1 + \tau(t)^2}{2} G^2(t) \bm{s}_\varphi(\bx_t, t)\bigg\} \diff t \nonumber \\
    & + \tau(t) G(t) \diff \Bar{\bm{w}}_t, ~ t \in [0, T], ~ \bx_T \sim p_T.
\end{align}

\paragraph{Classifier-free Guidance.}
Though the diffusion model has a solid theoretical interpretation, directly using its score functions to simulate the generative SDE often leads to sub-optimal performances, especially for class-conditioned generation. Assume $\bm{c}$ is a class label, and $\bm{s}_\varphi(\bx_t,t|\bm{c})$ is a conditional score network. The pioneering work \citet{ho2022classifierfreediffusionguidance} introduces the classifier-free guidance (CFG), which replaces the vanilla probability $q_t(\bx_t|\bm{c})$ with a new one $\Tilde{q}_t(\bx_t|\bm{c}) \coloneqq q_t(\bx_t)(\frac{q_t(\bx_t|\bm{c})}{q_t(\bx_t)})^{\omega}$. $q_t(\bx_t)$ represents the unconditional distribution which can be implemented by inputting an empty label $\bm{\varnothing}$ as $q_t(\bx_t|\bm{\varnothing})$. With this, the new score function turns to\begin{align}\label{eqn:cfg_score}
    \Tilde{\bm{s}}_{\varphi}(\bx_t, t|\bm{c}) \coloneqq \bm{s}_{\varphi}(\bx_t,t|\bm{\varnothing}) + \omega \big\{ \bm{s}_{\varphi}(\bx_t,t|\bm{c}) - \bm{s}_{\varphi}(\bx_t,t|\bm{\varnothing})\big\}
\end{align}
Such a guidance strategy has become a default setting for diffusion models, such as the Stable Diffusion series \citep{rombach2022high}. However, the CFG strategy has its limitations. First, CFG requires both a conditional and an unconditional score network, which either requires two separate models or is challenging to train with a single model. Second, the CFG is not available for tasks such as purely unconditional generation.

%% file: sec/4_connection.tex
\section{Self-Guidance}\label{sec:method}
\begin{figure*}
    \centering
    \includegraphics[width=1\linewidth]{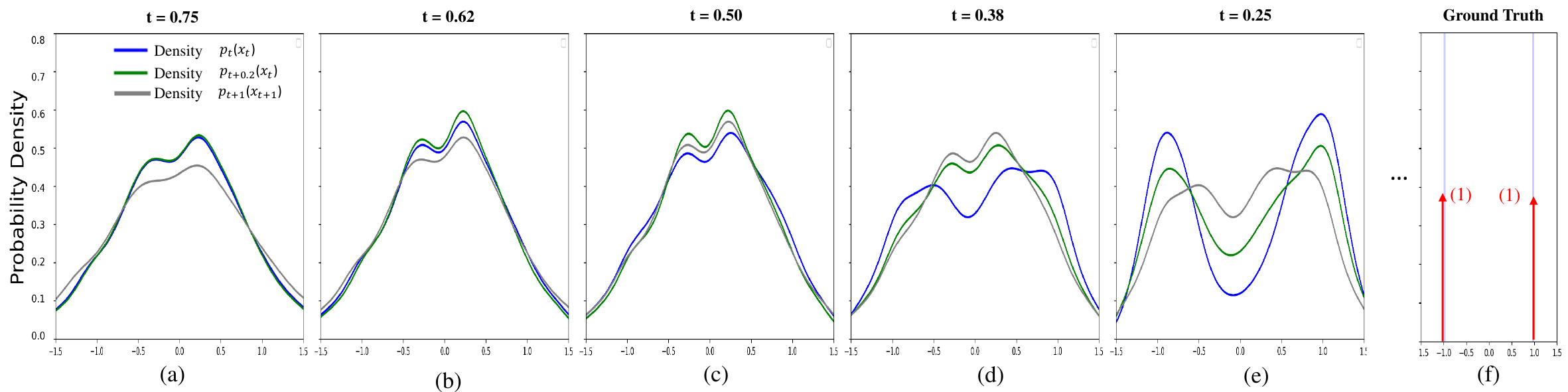}
    \caption{Comparison of SG and SG-prev under the two-mode distribution.  (a) - (e) are the results on different diffusion time from $t=0.75$ to $t=0.25$. The blue curve plots the distribution of generated samples at the current noise level $t$, the green curve plots the distribution of samples generated at the noisier level $t+\delta(t)$, while the gray curve plots the distribution of samples immediately generated at the immediately previous diffusion step.  (f) The ground truth two-mode distribution at $\pm 1$.}
    \label{fig:sgprev_explain}
\end{figure*}
\begin{figure*}
    \centering
    \includegraphics[width=1\linewidth]{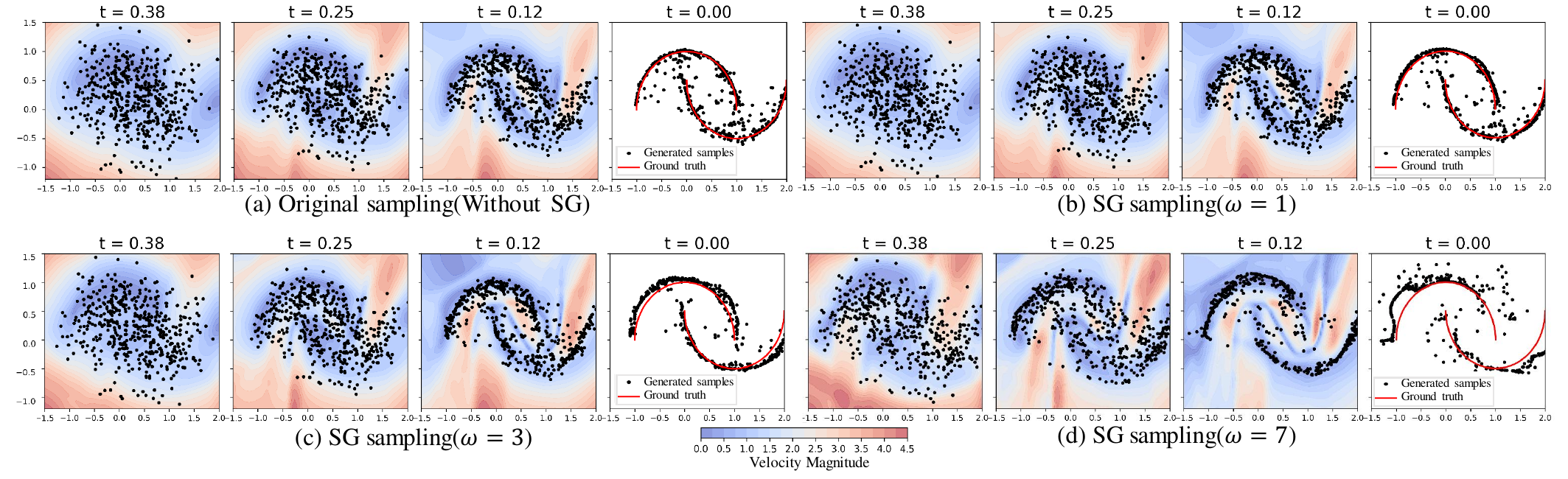}
    \caption{We train a flow-based diffusion model on a two-dimensional example with data sampled from a double-swirl pattern. We plot the distribution of the generated samples at the last four noise levels. The heat map in (a) shows the magnitude of predicted velocity without self-guidance, while the subfigures (b), (c), and (d) show those with Self-Guidance with various $\omega$ and noise levels. With a larger $\omega$, more artifact outliers are successfully removed. Meanwhile, a too large value of $\omega$ (e.g., $\omega=7$) could underweight (or, in other words, oversuppress) the sampling from the original density $p_t(\bx_t|c)$ that covers the swirl distribution. Thus, a suitable value of the guidance scale is necessary to balance the sampling coverage and the artifact suppression.}
    \label{fig:two_dim_distribution}
    \vspace{-5mm}
\end{figure*}

Figure~\ref{fig:two_point_distribution} demonstrates that using the ratio $\frac{p_t(\bx_t)}{p_{t+\delta(t)}(\bx_t)}$ helps suppress artifacts arising from the original diffused probability $p_t(\bx_t)$ at each noise level $t$. 

Formally, when generating the image with a condition $c$, the self-guided generative distribution at noise level $t$ can be formulated as discussed in Section~\ref{sec:intro},
\begin{align}\label{eqn:sg_density}
    p_{t}^{SG}(\bx_t|\bm{c},t) \propto & \, p_{t}(\bx_t|\bm{c}) 
    \bigg\{ \frac{p_t(\bx_t|\bm{c})}{p_{t+\delta(t)}(\bx_t|\bm{c})} \bigg\}^\omega.
\end{align}

This probability contains two parts. One is the original probability $p_t(x_t|c)$ that contains both the desired and artifact samples. The other is the ratio between probabilities of two diffusion times, and it is used to suppress the artifacts. The guidance scale $\omega$ acts as the combination weight.

Figure~\ref{fig:two_point_distribution} has already illustrated the influence of $\omega$ on the Self-Guidance sampling. 
Here we provide a more complex 2-dimensional example in Figure~\ref{fig:two_dim_distribution} to show the effects of Self-Guidance with various values of $\omega$. The ground truth data are sampled from a double-swirl pattern. By comparing the diffusion samplings with and without Self-Guidance, it demonstrates the ability of SG to remove the artifact outliers that do not reside on the swirls. With an increasing value of $\omega$, the outlier artifacts are suppressed more sharply, which eventually leads to the high-quality sampling of the ground truth distribution.  However, it is worth noting that when $\omega$ becomes too large (e.g., $\omega=7$), the distribution of generated samples could start drifting away from the swirls. This is not surprising because too large $\omega$ could underweight the original density $p_t(\bx_t|c)$ that aims to cover the samplings of the swirl data. This shows that a suitable guidance scale $\omega$ is necessary to balance between the sampling of true data density and the suppression of artifact outliers. In experiments, we will study the effect of its choices in the ablation.

Since diffusion models output the score instead of the diffused probability, one can transform Eq. \ref{eqn:sg_density}  into the score function by the relation of $\bm{s}(\bx_t|\bm{c},t)=\nabla_{x_t} \log p(\bx_t|c)$. Formally, we start with the log-density ratio between two nearby timesteps:
\begin{equation}
\log \frac{p_t(x_t|c)}{p_{t+\delta(t)}(x_t|c)} = \log p_t(x_t|c) - \log p_{t+\delta(t)}(x_t|c),
\label{eqn:sg_density}
\end{equation}
and take the gradient to $x$ to obtain the score difference:
\begin{align}
\nabla_x \log \frac{p_t(x_t|c)}{p_{t+\delta(t)}(x_t|c)} 
&= \nabla_x \log p_t(x_t|c) - \nabla_x \log p_{t+\delta(t)}(x_t|c) \nonumber \\
&= \bm{s}(x_t|c,t) - \bm{s}(x_t|c,t+\delta(t)).
\end{align}
Finally, the score with scale $\omega$ can be written as:
\begin{align}
\label{eqn:sg_score}
\bm{s}^{SG}(\bx_t|\bm{c},t) \coloneqq &\, \bm{s}(\bx_t|\bm{c},t) \nonumber \\
&+ \omega \Big\{ \bm{s}(\bx_t|\bm{c},t) - \bm{s}(\bx_t|\bm{c},t+\delta(t)) \Big\},
\end{align}

Fig.~\ref{fig:structure} illustrates how Self-Guidance is applied in practice at each denoising step during the reverse diffusion process. For a noisy latent $\bx_t$ at timestep $t$, the model simultaneously predicts the noise at $t$ and at a noisier timestep $t+\delta(t)$. These two predictions are then linearly combined as in Eq.~\ref{eqn:sg_score}, and the combined prediction is subsequently used to denoise $\bx_t$ and obtain the latent $\bx_{t-1}$ for the next denoising step.
This can be easily implemented by inference with the same model with two different timesteps within a batch. This means SG does not require an additional network or specific training tricks as in other guidance algorithms, but only uses the trained model itself. That is why we called the method Self-Guidance.

There are various choices of the shifted noise scale $\delta(t)$. In this paper, we consider a constant shift scale that is fixed independent of diffusion time $t$, and a dynamic one $\delta(t)=t/\sigma$ that becomes smaller as $t$ approaches $0$. The latter allows a stronger Self-Guidance at the beginning of the diffusion sampling process. Moreover, we can also simply reuse the output distribution $\bm{s}(\bx_{t+1}|\bm{c},t+1)$ from the previous diffusion step to approximate the SG guidance term, as in Eq.~\ref{eqn:sg_prev_score}
\begin{align}
    \label{eqn:sg_prev_score}
    \bm{s}^{SG\_prev}(\bx_t|\bm{c},t) \coloneqq & \, \bm{s}(\bx_t|\bm{c},t) \nonumber \\
    & + \omega \bigg\{ \bm{s}(\bx_t|\bm{c},t) - \bm{s}(\bx_{t+1}|\bm{c},t+1) \bigg\}.
\end{align}
This avoids doubling inference time as it does not require an additional forward pass at each diffusion step. We call this approximation method SG-prev. 

In Figure~\ref{fig:sgprev_explain}, we present a comparison between SG and SG-prev on the same one-dimensional toy example as in Figure~\ref{fig:two_point_distribution}. We find that the so-called SG-prev method is more efficient while bringing additional positive effects.

During image denoising, the early steps are mostly noise, and the image gradually becomes clear in the final steps. This is reflected by the probability density $p_{t}(x_{t})$ in Figure~\ref{fig:sgprev_explain}: at high-noise levels, the distribution is nearly Gaussian, so changes between steps are small; at low-noise levels, the distribution varies significantly, making the previous diffusion step probability $p_{t+1}(x_{t+1})$ (gray curve) near $x=0$ higher than $p_{t+\delta(t)}(x_t)$ at the slightly noisier level $t+\delta(t)$ (green curve), which leads to stronger suppression of artifacts.

{
\setlength{\floatsep}{0pt}
\setlength{\textfloatsep}{0pt}
\begin{algorithm}[h]
\small
\SetAlgoLined
\KwIn{Conditional DM $d_{cod(\theta)}$, Unconditional DM $d_{uncod(\theta)}$, Perturbed DM $\hat{d}_{cod(\theta)}$, guidance scales $\omega_{CFG},\omega_{PAG},\omega_{SG}$, shift scales $\delta(t)$.}
\For{ $t$ in \texttt{timesteps}}{
$\bm{s}(\bx_t|\bm{c},t) = d_{cod(\theta)}(\bx_t, \bm{c}, t)$ \\
\If{Classifier-free guidance}{
$\bm{s}(\bx_t|t) = d_{uncod(\theta)}(\bx_t, t)$ \\
}
\If{Perturbed-attention guidance}{
$\bm{\hat{s}}(\bx_t|\bm{c},t) = \hat{d}_{cod(\theta)}(\bx_t, \bm{c}, t)$ \\
}
\If{Self-Guidance}{
  $\bm{s}(\bx_t|\bm{c}, t+\delta(t))$ \\
    ~~~~$= d_{\text{cod}(\theta)}(\bx_t, \bm{c}, t+\delta(t))$
    ~~\text{if SG} \\
    ~~~~$\approx d_{\text{cod}(\theta)}(\bx_{t+1}, \bm{c}, t+1)$ ~~\text{if SG-prev}
}
    $\bm{s}^*(\bx_t|\bm{c},t) = \bm{s}(\bx_t|t)$\\ ~~~~~~~~~~~~~~~~~~$+ \omega_{CFG} * (\bm{s}(\bx_t|\bm{c},t) - \bm{s}(\bx_t|t))$ \\
       ~~~~~~~~~~~~~~~~~ $+ \omega_{PAG} * (\bm{s}(\bx_t|\bm{c}, t) - \bm{\hat{s}}(\bx_t|\bm{c},t))$ \\
        ~~~~~~~~~~~~~~~~~ $+ \omega_{SG} * (\bm{s}(\bx_t|\bm{c}, t) - \bm{s}(\bx_t|\bm{c},t+\delta(t)))$

    $\bx_{t-1}$ = \texttt{scheduler.step}($\bm{s}^*(\bx_t|\bm{c},t), t, \bx_t$) \\
}
\Return{$\bx_0$}
\caption{Diffusion model inference with Self-Guidance and other guidance}
\label{alg:self_guidance}
\end{algorithm}
}

We detail how SG and SG-prev are used in inference in Algorithm \ref{alg:self_guidance}, as well as if combined with other diffusion guidance methods, such as CFG and PAG.
For a single step at the noise level $t$, the diffusion model applies different guidance individually by forwarding the model with the corresponding settings: unconditional prediction in CFG, perturbing the attention in PAG, a shifted timestep $t+\delta(t)$ in SG, and the previous timesteps' output in SG-prev. These correct terms provide orthogonal improvements. Combined with their corresponding weight factors $\omega_{CFG},\omega_{PAG},\omega_{SG}$, the overall prediction is optimized to achieve higher consistency and control in the quality of the generation.

\paragraph{Some discussions about the motivation of SG}
Besides the empirical motivation, Self-guidance also shares solid theoretical backgrounds. In this part, we show that suppressing artifacts with $\frac{p_t(\bx_t|\bm{c})}{p_{t+\delta(t)}(\bx_t|\bm{c})}$ can be motivated from the heat equation. Specifically, under Gaussian smoothing with kernel $\mathcal{N}(0, \sigma^2)$, the diffused density evolves as:
\begin{align}
p_{t+\delta(t)}(x) 
&= \int p_t(y) \, \mathcal{N}(x; y, \sigma^2\delta(t))\, dy \notag\\
&= (p_t * \mathcal{N}(0, \sigma^2))(x).
\end{align}
Differentiating with respect to $t$ yields the classical heat equation:
\begin{equation}
\partial_t p_t(x) = \frac{1}{2} g(t)^2 \Delta_x p_t(x),
\end{equation}
where $g(t)$ denotes the noise scaling function and $\Delta_x$ is the Laplacian operator. Now, assume that an artifact resides at a location $x_{\text{artifact}}$ in a low-density region of $p_t(x)$, i.e., a “valley.” Under the multivariate second-derivative test, the Hessian at this point is positive definite, implying
\begin{equation}
\Delta_x p_t(x_{\text{artifact}}) > 0.
\end{equation}
Thus, the heat equation implies:
\begin{equation}
\partial_t p_t(x_{\text{artifact}}) > 0 \quad \Rightarrow \quad p_{t+\delta(t)}(x_{\text{artifact}}) > p_t(x_{\text{artifact}})
\end{equation}
when $\delta(t)$ is sufficiently small.

In other words, forward diffusion causes density to flow into low-density artifact regions, increasing their mass. Therefore, comparing $p_t$ to $p_{t+\delta(t)}$ (via their ratio) effectively downweights these artifact regions and promotes sampling from the true data manifold.

%% file: sec/5_experiment.tex
\section{Experiments} \label{sec:exp}

\begin{table*}[!ht]
    \small
    \centering
    \caption{Quantitative results of different Stable Diffusion models and Flux with various guidance. FID, Clip Score, and Aesthetic Score are evaluated on the MS-COCO 2017 validation subset. Bolded values highlight the best performance.}
    \setlength{\tabcolsep}{1pt}
    \begin{tabular}{llcccccccc} 
        \toprule
        \multirow{2}{*}{Model}&\multirow{2}{*}{Approch}& \multirow{2}{*}{FID$\downarrow$} & \multicolumn{5}{c}{HPS v2.1} & \multirow{2}{*}{Clip Score$\uparrow$}& \multirow{2}{*}{Aesthetic Score$\uparrow$} \\ 
         && & concept-art$\uparrow$ & photo$\uparrow$ & anime$\uparrow$ & paintings$\uparrow$ & average$\uparrow$ & & \\ 
        \midrule
        \multirow{8}{*}{\centering SD 1.4\cite{rombach2022high}}&baseline& 50.76& 21.22 & 21.24 & 21.37 & 20.52 & 21.09 & 0.2695 & 5.0025 \\
 &SG& 42.73& 22.13& 22.23& 21.84& 21.73& 21.98& 0.2781&5.0766\\ 
        &CFG\cite{ho2022classifierfreediffusionguidance}& 27.07& 23.61& 24.91& 23.60& 24.83& 24.24& 0.3095& 5.4427\\ 
        &CFG+PAG\cite{ahn2024selfrectifyingdiffusionsamplingperturbedattention}& 27.37& 23.61& 24.87& 24.86& 23.73& 24.26& 0.3093& \textbf{5.4448}\\
 & CFG+SG& 26.67& \textbf{24.95}& 25.17& 24.13& 24.12& 24.59& 0.3094&5.4201\\
 & CFG+SG-prev& 23.70& 24.90& 23.73& 23.87& \textbf{25.10}& 24.39& 0.3110&5.2954\\   
        &CFG+PAG+SG& 26.75
& 24.12& \textbf{25.28}& \textbf{24.99}& 24.20& \textbf{24.65}& 0.3094& 5.4351\\
 & CFG+PAG+SG-prev& \textbf{23.58}& 23.94& 25.17& 24.91& 23.75& 24.44& \textbf{0.3115}&5.3444\\   
        \midrule
        \multirow{8}{*}{\centering SD 3\cite{esser2024scalingrectifiedflowtransformers}}&baseline& 51.47& 21.17 & 18.91 & 21.11 & 21.62 & 20.70& 0.2762 & 4.9628\\
 &SG& 48.82& 22.01& 19.49& 20.79& 21.47& 20.83& 0.2964&4.8888\\  
        &CFG\cite{ho2022classifierfreediffusionguidance}& 27.00& 30.26& 28.29& 30.81& 30.62& 30.00&0.3169 &  5.3915\\   
        &CFG+PAG\cite{ahn2024selfrectifyingdiffusionsamplingperturbedattention}& 27.34& 30.26& 28.45& 30.78& 30.73& 30.05& 0.3164&5.3978 \\
 & CFG+SG& 25.21& \textbf{30.67}& 28.49& \textbf{31.10}& \textbf{31.03}& 30.32& 0.3163&\textbf{5.4431}\\
 & CFG+SG-prev& 25.15& 29.94&  27.57& 30.33& 30.17& 29.50& 0.3217&5.3260\\   
        &CFG+PAG+SG& 25.28& 30.65& \textbf{28.72}& 31.09& 31.02& \textbf{30.37}& 0.3162&5.4424 \\
 & CFG+PAG+SG-prev& \textbf{24.83}& 29.92& 27.79& 30.34& 30.26& 29.58& \textbf{0.3219}&5.3238\\  
        \midrule
        \multirow{5}{*}{\centering SD 3.5\cite{stability2023stable_diffusion}}&baseline& 39.69& 22.06& 22.98& 23.87& 22.03& 22.74& 0.3020& 5.3068\\
 &SG& 38.40& 23.74& 24.74& 25.08& 23.39& 24.24& 0.3018&5.3986\\   
        &CFG\cite{ho2022classifierfreediffusionguidance}& 23.86& 31.67& 29.48& \textbf{32.65}& 31.43& 31.31&0.3129 &5.6152 \\ 
        &CFG+SG& 22.70& \textbf{31.74}& \textbf{29.50}& 32.57&\textbf{31.52} & \textbf{31.33}& 0.3132& \textbf{5.6532}\\
 & CFG+SG-prev& \textbf{20.74}& 31.41& 28.93& 32.07& 31.24& 30.91& \textbf{0.3162}&5.6013\\  
        \midrule
        \multirow{3}{*}{\centering Flux.1\cite{blackforestlabs} }&CFG\cite{ho2022classifierfreediffusionguidance}& 29.74& 31.26& 29.87& 32.80& 31.78& 31.43 &0.3080&5.7527 \\  
        &CFG+SG& 28.59& 31.32& \textbf{30.02}& 32.89& 31.81& 31.51& 0.3084&\textbf{5.7845} \\ 
 & CFG+SG-prev& \textbf{25.54}& \textbf{31.54}& 29.71& \textbf{32.92}& \textbf{32.06}& \textbf{31.56} & \textbf{0.3129}&5.7391\\
         \bottomrule                                  
    \end{tabular}
    \vspace{-3mm}
    \label{tab:different guidance compare}
\end{table*}

\begin{table*}
    \small
    \setlength{\tabcolsep}{1pt}
    \centering
    \caption{Quantitative results of SG Performances in Specific Generative Domains compared with CFG baseline Flux.1 \cite{blackforestlabs}.} 
    \begin{tabular}{ccccccccc}
    \toprule
    \multirow{2}{*}{\textbf{Methods}} & \multicolumn{2}{c}{\textbf{Hand Generation}} & \multicolumn{2}{c}{\textbf{Face Generation}} & \textbf{Text-Image Alignment} & \multicolumn{3}{c}{\textbf{Generation Quality}} \\ \cline{2-9} 
      & FID-H $\downarrow$ & Hand-Conf $\uparrow$& FID-F $\downarrow$& FaceScore $\uparrow$& CLIP Score $\uparrow$& FID $\downarrow$& Pick Score$\uparrow$& ImageReward$\uparrow$\\ \midrule
    Flux.1\cite{blackforestlabs}& 67.0163 & 0.8902 & 34.1976 &  4.5448& 0.3080 &  29.74& 22.9802&  1.0971\\ 
     +SG& \textbf{66.4360} & \textbf{0.9283} & \textbf{34.0561} &  \textbf{4.6634}& \textbf{0.3084} &  \textbf{28.59}& \textbf{22.9960}&  \textbf{1.1046}\\ 
 \bottomrule
    \end{tabular}
    \label{tab:generation_metrics}
    \vspace{-3mm}
\end{table*}
\begin{figure*}
    \centering
    \includegraphics[width=1\linewidth]{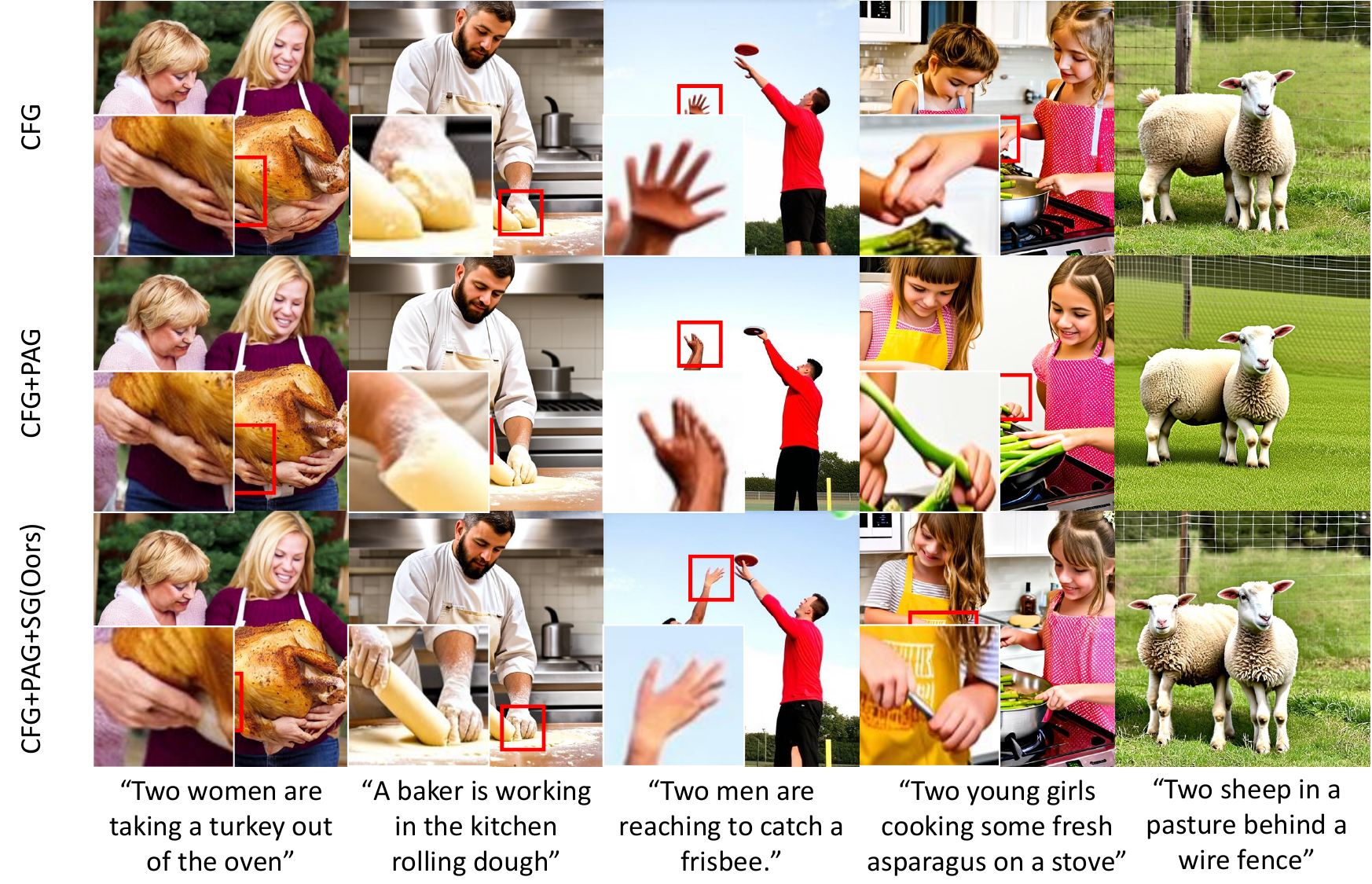}
    \caption{Qualitative comparison between CFG, CFG+PAG and CFG + PAG+SG. The red box in the figure indicates where the baseline (SD1.4 and SD3) is poorly generated, and the zoomed-in image is shown in the lower left corner.}  
    \label{fig:different_guidance_combine}
\end{figure*}

\subsection{Implementation Details}
\paragraph{Models}
We apply our Self-Guidance on current state-of-the-art text-to-image and text-to-video models. For text-to-image, we consider both a classic diffusion model, Stable Diffusion 1.4 \cite{rombach2022high}, and current state-of-the-art flow-based models, Stable-Diffusion 3 series \citep{esser2024scalingrectifiedflowtransformers,stability2023stable_diffusion} and Flux.1-dev\cite{blackforestlabs}. For text-to-video, we consider CogVideoX-5B \cite{yang2024cogvideoxtexttovideodiffusionmodels}, an open-source video generation model trained with flow matching.

To facilitate a fair comparison, we consistently evaluate the model with the same number of steps in inference: DDIM Solver with 50 steps for SD 1.4, Euler with 28 steps for SD 3 series, and FLUX. The guidance scales for CFG and PAG are kept the same as the original settings in the corresponding papers. We will elaborate in Appendix \ref{app:c3}.

\paragraph{Evaluation Metrics}
We employ Fréchet Inception Distance (FID) \cite{heusel2018ganstrainedtimescaleupdate}, Human Preference Score (HPS v2.1) \cite{wu2023humanpreferencescorev2}, CLIP Score \cite{taited2023CLIPScore}, and Aesthetic Score\cite{improvedaestheticpredictor} to comprehensively assess the quality of the image generation. For video generation, we employ VBench\cite{huang2023vbenchcomprehensivebenchmarksuite} to decompose video generation quality into multiple well-defined dimensions.

In addition, to address and analyze the performance on some long-standing problems in image generation, we introduce some specific evaluation metrics, including Hand-Conf \citep{heusel2018ganstrainedtimescaleupdate,parmar2022aliasedresizingsurprisingsubtleties}, Hand-FID (FID-H) \cite{heusel2018ganstrainedtimescaleupdate} in hand generation, FaceScore\cite{liao2024facescorebenchmarkingenhancingface}, Face-FID (FID-F) \cite{heusel2018ganstrainedtimescaleupdate} in face generation, Pick Score\cite{kirstain2023pickapicopendatasetuser}, ImageReward \cite{xu2023imagerewardlearningevaluatinghuman} in high-quality generation. We will elaborate on them in detail when used later.

\subsection{Quantitative Comparisons }\label{sec:5.2}
\subsubsection{Text-to-Image Generation}\label{sec:5.2.1}

Table \ref{tab:different guidance compare} shows the effect of using SG itself on the three SD models. 
First of all, by incorporating SG itself on SD 1.4, we achieve an impressive 8-point (50.7602$\to$42.7362) improvement in the FID. Similar improvements appear in other metrics and models as well. This indicates that SG itself is able to improve the quality of generated images.

SG is specifically designed to address the artifact issues in CFG and PAG. As shown in Table \ref{tab:different guidance compare}, incorporating SG with CFG leads to consistent improvements in HPS across all models, demonstrating that SG better aligns with human preferences. Moreover, our approach remains competitive with PAG in both CLIP and Aesthetic Score.

Then, we find that combining SG with CFG and PAG will have the best performance over using them alone. Figure~\ref {fig:different_guidance_combine} shows the comparison between CFG, CFG+PAG, and CFG + PAG+SG. The results show that adding SG not only enhances fine-grained details, prompt alignment, and error correction but also achieves new SoTA better performance across multiple metrics (24.65 HPS v2.1 for SD 1.4, 25.28 FID for SD 3, 0.3132 CLIP Score in SD 3.5, and 5.7845 Aesthetic Score in Flux).

In addition, SG-prev outperformed both compared baselines and the SG model, particularly in terms of FID and CLIP Score. This is achieved by applying SG-prev after the diffusion time (not step) $t$ decreases below 500, according to Figure~\ref{fig:sgprev_explain}. This demonstrates that the SG-prev successfully boosts the diffusion models without incurring any additional cost.

\subsubsection{Text-to-Video Generation}\label{sec:5.2.2}

\begin{figure*}
    \centering
    \includegraphics[width=1\linewidth]{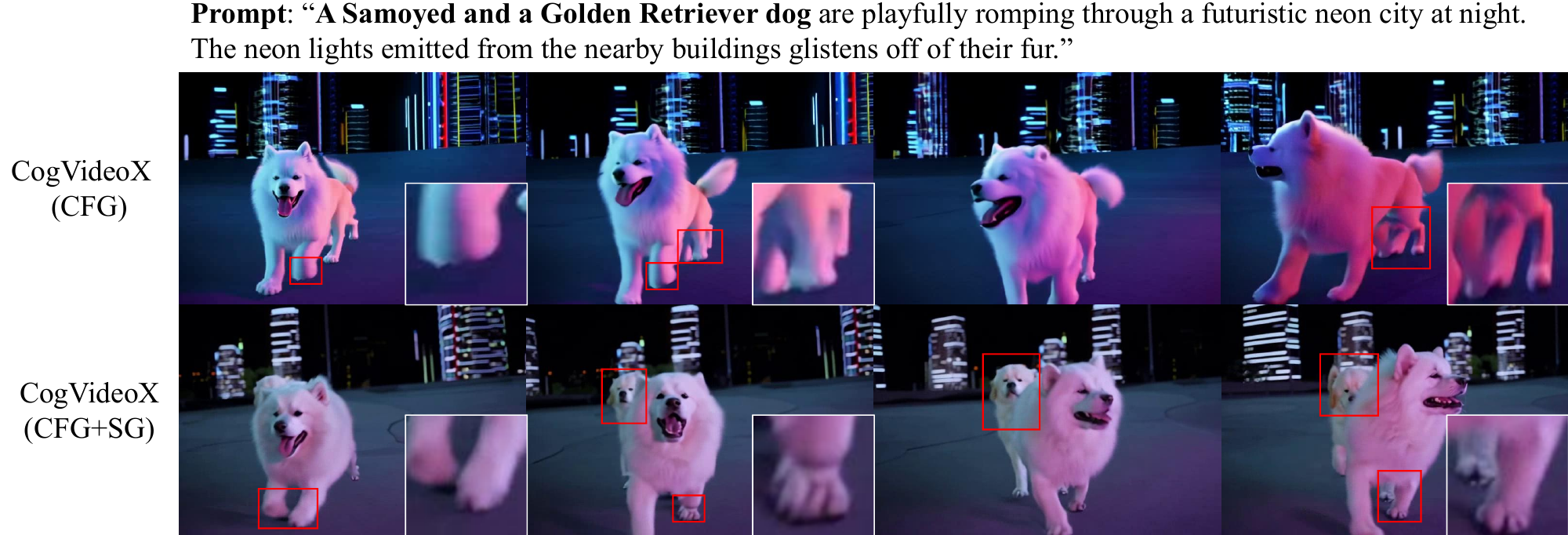}
    \caption{Qualitative comparison between CFG and CFG+SG in text-to-video model CogVideoX-5B\cite{yang2024cogvideoxtexttovideodiffusionmodels}. The red box in the figure indicates where the baseline(CFG) is poorly generated, and the zoomed-in image is shown in the lower left corner. Adding SG to CFG effectively eliminates artifacts (Broken arms and legs and misplaced limbs) in video generation. SG also enhances the consistency between prompt and video(Number of dogs in the video).}
    \label{fig:video_example}
\end{figure*}

Table \ref{tab:video_benchmark} provides quantitative evaluation results across four key dimensions in VBench \cite{huang2023vbenchcomprehensivebenchmarksuite}. The results indicate that adding SG effectively improves the metrics in human action (0.8320$\to$0.8450), and multiple objects (0.4619$\to$0.4703). SG-prev also outperforms the baseline across all four metrics under the same sampling time, but slightly lags behind SG that doubles the sampling time. Additionally, We provide more video examples in Appendix \ref{app:A.2}. After having a close look at the generated examples in Figure~\ref{fig:video_example}, we find that SG also effectively eliminates artifact problems (broken arms and legs or misplaced limbs) in video generation tasks and enhances text-video consistency. Due to this comparison, we may conclude that SG is also effective in improving video generation quality.
.

\subsection{Guidance Performances in Specific Domains}\label{sec:5.3} 
Through our experiments and analysis (e.g., Figure~\ref{fig:specific_task_generation}), we observed that SG is particularly effective when handling some long-standing problems in image generation.
More specifically,
SG can effectively generate hands with an exact correct number of fingers with natural shapes (Part 1), and remove artifacts like redundant arms or hands (Part 2). It also enhances Flux's comprehension of textual input (Part 4), effectively eliminating extraneous objects. This allows the main subject of the image to stand out while the background is appropriately blurred(Part 3). 

To quantitatively reveal the effectiveness of SG in these aspects, we propose some domain-specific metrics and conduct a comparison with Flux.1 \cite{blackforestlabs}, the state-of-the-art text-to-image model. Here we introduce the metrics in detail.
\paragraph{Hand Generation}
To evaluate the quality of the hands in the generated images, we propose two metrics, the Fréchet Inception Distance for hands (FID-H) and the HAND-CONF. FID-H selects 5,000 human-hand-related prompts from the coco validation set, and calculates the distance between the generated image and the real images in COCO. To compute HAND-CONF, we detect hands on the generated images with a pretrained hand detector (i.e. Mediapip \cite{zhang2020mediapipehandsondevicerealtime}, and average the confidence scores on 5,000 prompts in the HandCaption-58k dataset \cite{NagolincHandCaptions}.

\paragraph{Face Generation}
Fréchet Inception Distance for Faces (FID-F) is calculated with 5,000 face-related prompts and images in COCO val. Regarding FaceScore, we use a face quality-focused reward model \cite{liao2024facescorebenchmarkingenhancingface} to score 10,000 images generated with prompts from the HumanCaption-10M dataset \cite{OpenFaceCQUPTHumanCaption10M}.

Moreover, to assess text-image consistency, we measured the CLIP Score. FID, PickScore \cite{kirstain2023pickapicopendatasetuser} and ImageReward \cite{xu2023imagerewardlearningevaluatinghuman} are also listed here for general image quality assessments.

As shown in Table \ref{tab:generation_metrics}, SG achieves significant results in fine-grained generation tasks, such as hand and face generation, with improvements in FID-H (67.0163 to 66.4360) and FID-F (34.1976 to 34.0561). Additionally, the increase in PickScore and ImageReward indicates that our SG is more adept at producing higher-quality samples, thus minimizing generation errors, as illustrated in Part 4 of Figure \ref{fig:specific_task_generation}.

In summary, our SG improves diffusion and flow-based models in multiple aspects, generating high-quality images.
\begin{table*}
    \small
    \centering
    \caption{Quantitative results of CFG based text-to-video generation model CogVideoX with our SG and SG-prev. All scores are evaluated on the Standard Prompt Suite of VBench\cite{huang2023vbenchcomprehensivebenchmarksuite}. Bolded values highlight the best performance for each metric.}
    \begin{tabular}{ccccc}
    \toprule    
         Model&  Human Action$\uparrow$ &Overall Consistency$\uparrow$
&  Multiple Objects$\uparrow$&  Appearance Style$\uparrow$\\
         \midrule
         CogVideoX-5B\cite{yang2024cogvideoxtexttovideodiffusionmodels}&  0.8320 &0.2526
&  0.4619&  0.2331\\
         + SG&  \textbf{0.8450} &\textbf{0.2553}&  0.4703&  \textbf{0.2339}\\
 +SG-prev& 0.8376& 0.2530& \textbf{0.4774}&0.2333\\
          \bottomrule
    \end{tabular}
    
    \label{tab:video_benchmark}
\end{table*}
\begin{figure}
    \centering
    \includegraphics[width=1\linewidth]{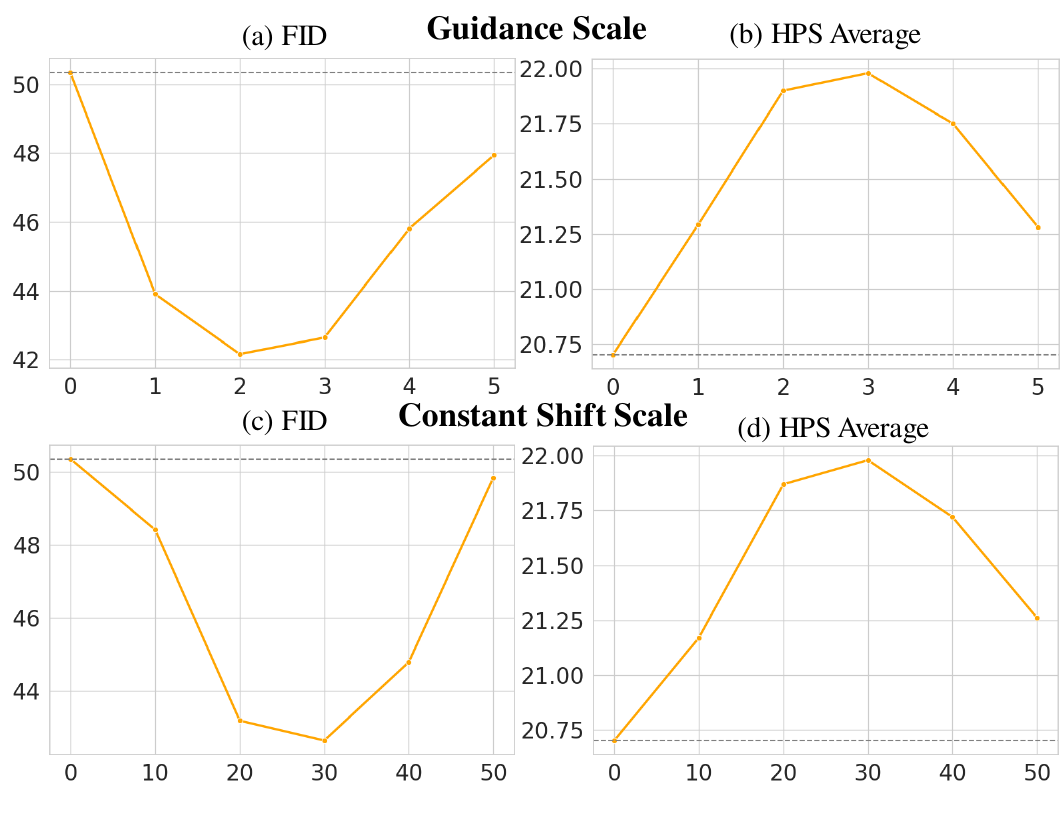}
    \caption{Ablation study of the Guidance Scale and Shift Scale.}
    \label{fig:ablation different scales}
    \vspace{-5mm}
\end{figure}

\subsection{Ablation Studies}

\subsubsection{Guidance Scale}
We conducted an experimental investigation into the effect of the guidance scale on the performance of Self-Guidance. Using Stable Diffusion 1.4 \cite{rombach2022high}, we sampled 5000 images with guidance scales ranging from 0.0 to 5.0 in 1.0 intervals. We measure the FID \cite{heusel2018ganstrainedtimescaleupdate} and HPS v2.1 \cite{wu2023humanpreferencescorev2} for these images. The results, as shown in Figure \ref{fig:ablation different scales}, indicate that Self-Guidance achieves the best FID (42.23) at a guidance scale of 2.0, and the highest HPS (21.98) at 3.0. 

\subsubsection{Shift Scale}
\paragraph{Constant Shift Scale}
The shift scale $\delta(t)$ represents the extent to which the noise level deviates from the current level. A simple way is to set it to a constant independent of $t$. The larger the shift scale is, the more it can contrast with the current noise level to suppress the artifacts. However, too large a shift may damage the desired patterns that should be preserved in the current sampling density, and thus a suitable shift scale should be adopted.
Figure \ref{fig:ablation different scales} shows how the FID and HPS score change as the shift scale increases from 0 to 50 (within 1,000 total training timesteps). According to the results, SG achieves the lowest FID and highest HPS score with the shift scale set to 30.

\paragraph{Dynamic Shift Scale}
One can also dynamically adjust $\delta(t)$ during the reverse diffusion process. At higher noise levels, a large shift of $\delta(t)$ can be adopted, as noisy artifacts can be rapidly removed at the early stage of diffusion sampling. 
Thus, we introduce $\delta(t)$ as a linear function of time ($\delta(t) = t / \sigma$), allowing stronger guidance at higher noise levels while reducing it at lower levels. More experiments are shown in Appendix \ref{app:A.4}. We find that dynamic shift scaling improves quality, avoiding blurred images or noise issues caused by excessive guidance as the noise level decreases to zero.

\subsubsection{Hyperparameter Selection Strategy}
Based on our experiments, setting the guidance scale ($\omega$) to 3 generally yields the best results across different models. However, slight tuning may still be necessary depending on the specific model.  
If the generated images appear overly blurry after setting $\omega$ = 3, increasing $\omega$ moderately (e.g., to 3.5–4) can help enhance sharpness. Conversely, if excessive noise is observed, decreasing $\omega$ (e.g., to 1–2) is recommended.

For the shift scale ($\delta(t)$), we use a default value equal to 1\% of the model’s total diffusion time (e.g., $\delta(t)$ = 10 if diffusion time = 1000). The tuning strategy for $\delta(t)$ follows the same principle as for $\omega$: increase $\delta(t)$ (e.g., to 10-20) if images are too blurry, or decrease it (e.g., to 1–10) if observe excessive noise.

That said, we recommend using SG-prev as a more practical alternative, since it can be viewed as a form of dynamic shift scale. With SG-prev, only the guidance scale needs to be tuned, simplifying hyperparameter selection.

%% file: sec/6_conclusion.tex
\section{Conclusions}

This paper suggests that the reason for artifacts in image generation is incompetent denoising at each step, and proposes Self-Guidance (SG)—an inference-time strategy that modulates the output distribution using the difference between distributions at the current step $t$ and a noisier step $t+\delta(t)$. SG is architecture-agnostic, training-free, and compatible with existing guidance methods. Extensive experiments show that SG consistently improves text-to-image and text-to-video generation, especially in challenging cases like generating realistic human hands and bodies. To reduce SG’s two forward passes per sampling step, we introduce SG-prev, which approximates SG by directly reusing the output from the immediately previous timestep. As a result, SG-prev introduces no additional computational cost.

%% file: sec/X_suppl.tex
\appendix
\subsection{Additional results}

\subsubsection{More results of SG combining with CFG and PAG} \label{app:A.1}
Figure 

\begin{figure*}
    \centering
    \includegraphics[width=1\linewidth]{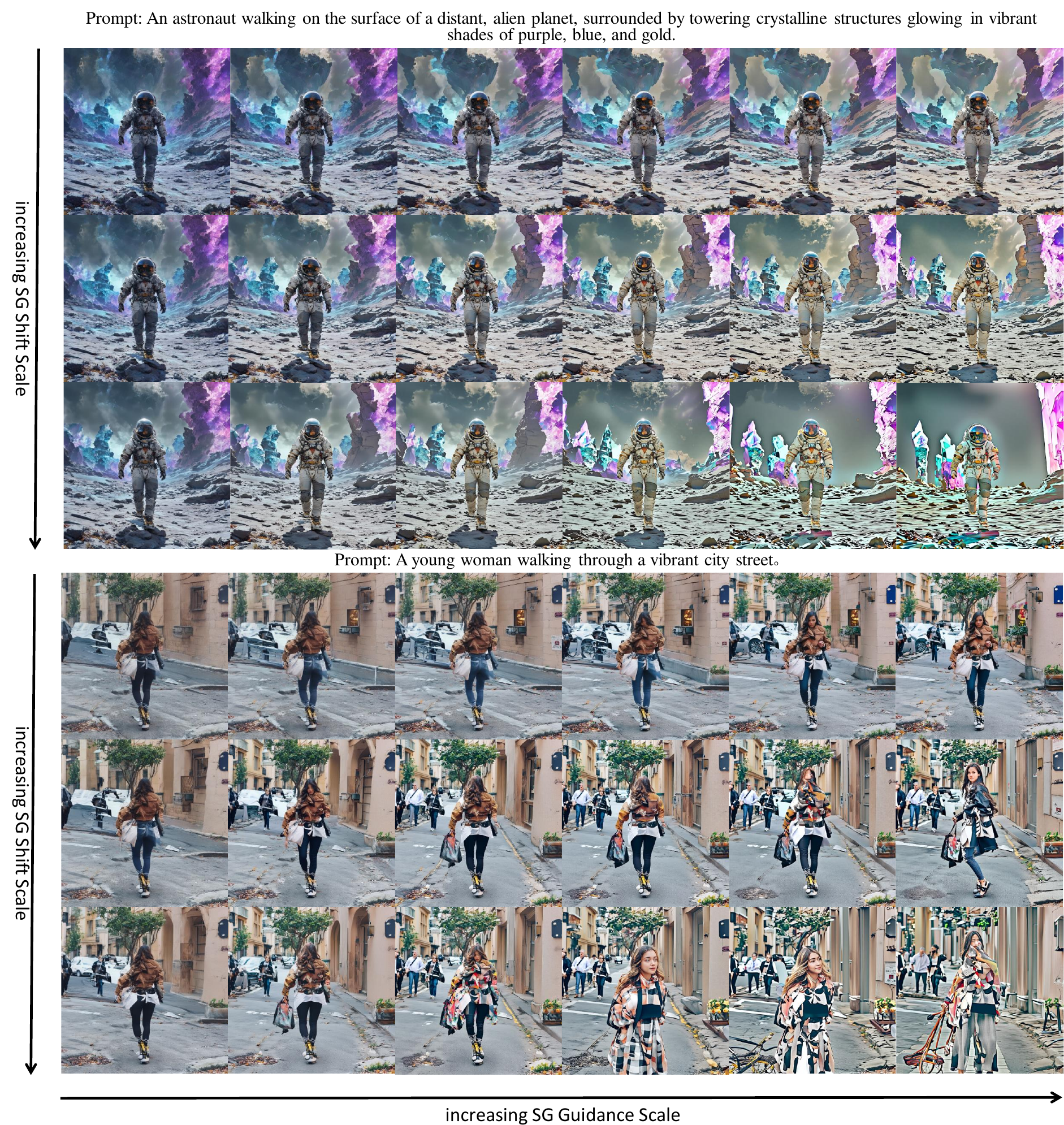}
    \caption{SG results of different guidance scales and shift scales on Stable Diffusion 3-medium.}
    \label{fig:guidance example}
\end{figure*}

\begin{figure*}
    \centering
    \includegraphics[width=1\linewidth]{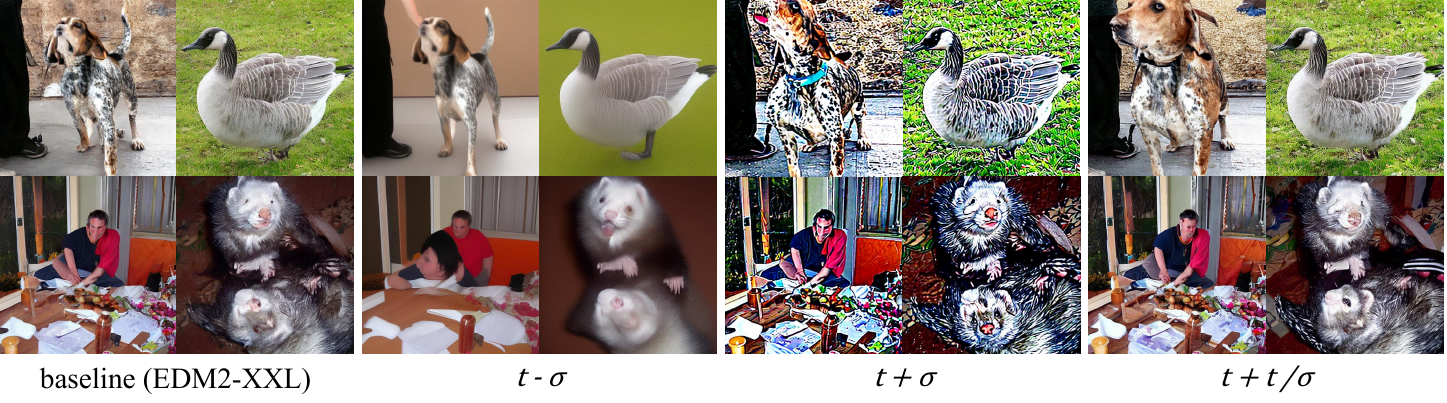}
    \caption{Ablation study of the dynamic shift scale on EDM2-XXL\cite{karras2024analyzingimprovingtrainingdynamics}. }
    \label{fig:ablation_dynamic_shift}
\end{figure*}

\subsubsection{More results of Text-to-Video Generation} \label{app:A.2}
\begin{figure*}
    \centering
    \includegraphics[width=1\linewidth]{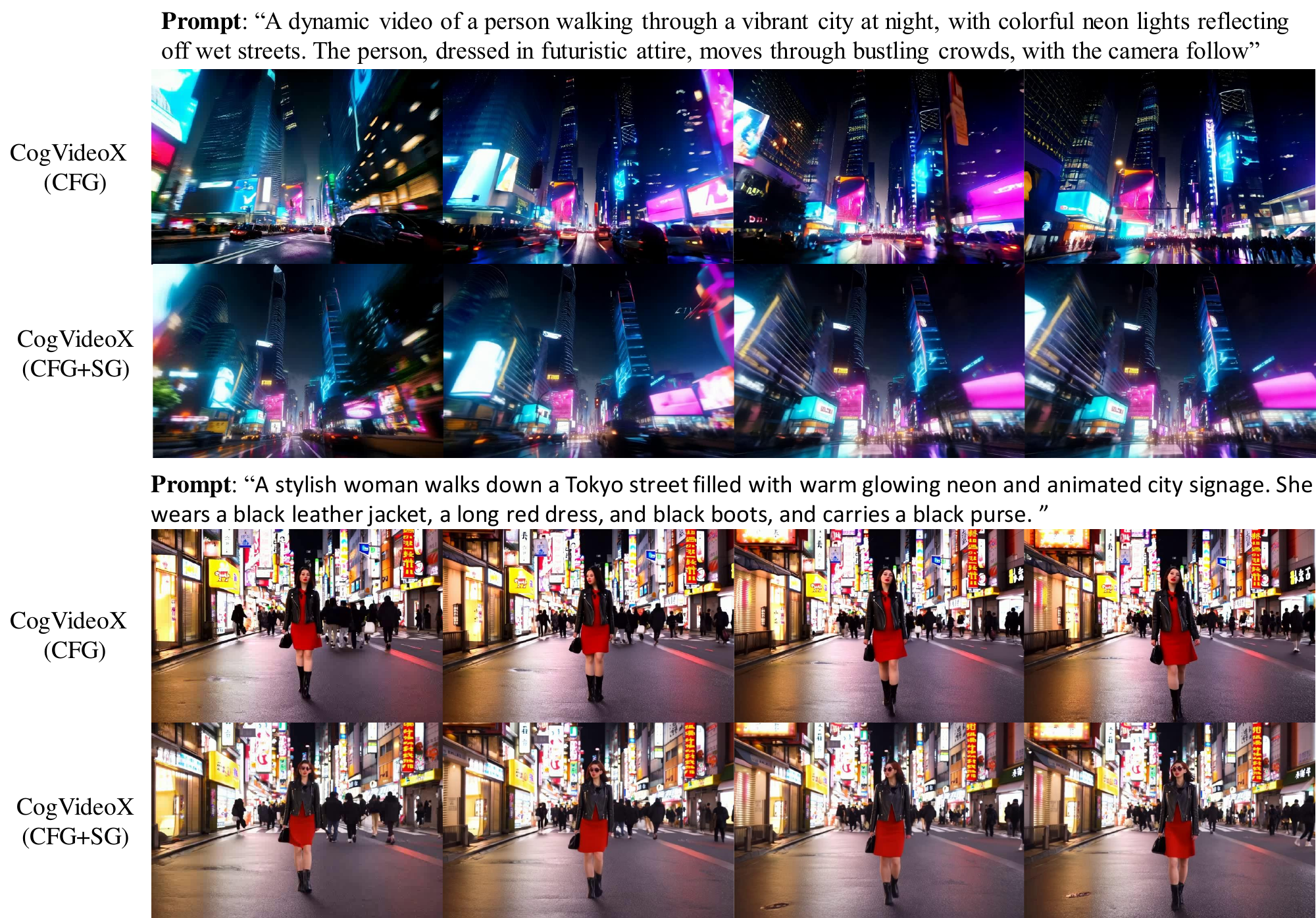}
    \caption{Additional results for using SG on Text-to-Video models, similar to Figure \ref{fig:video_example}}
    \label{fig:video example}
\end{figure*}

Figure \ref{fig:video example} shows more results of our SG on Test-to-Video models. For complete video examples, see the videos folder of the Appendix.

\begin{figure*}
    \centering
    \includegraphics[width=1\linewidth]{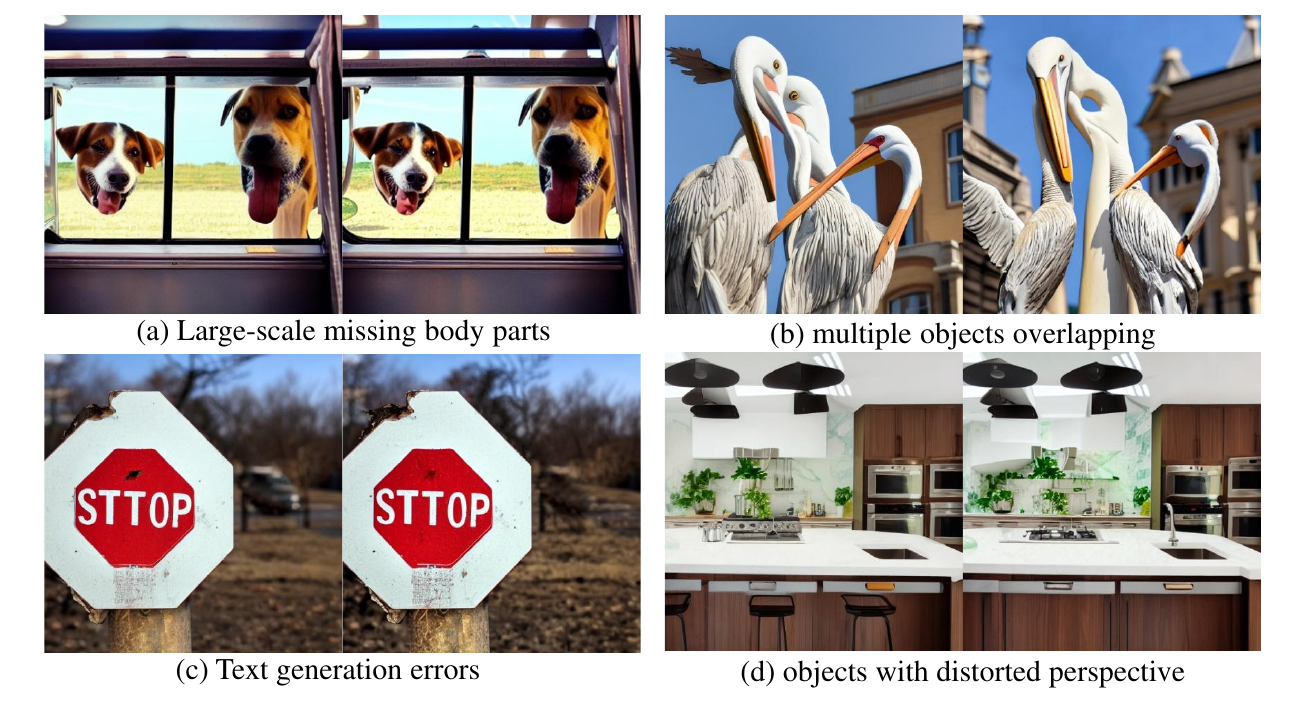}
    \caption{Examples illustrating the limitations of SG in making global structural corrections. Left is the image generated by SD 1.4, right is SD1.4 + SG.}
    \label{fig:limitations}
\end{figure*}

\subsubsection{More results of ablations of guidance scale and shift scale} \label{app:A.3}
Figure \ref{fig:guidance example} shows the results of different guidance scales and shift scales on Stable Diffusion 3-medium. We can find that with the increase of the guidance scale and shift scale, our SG can significantly improve the guidance performance.
\subsubsection{More results of ablations of dynamic shift scale} \label{app:A.4}
Figure \ref{fig:ablation_dynamic_shift} shows the result of the dynamic shift scale on the EDM2-XXL model. $\sigma$ is a fixed value. We let the shift scale $\delta(t)$ be $-\sigma$, $\sigma$, and $t/\sigma$. The results show that using fixed $\delta(t)$ caused either blurred images or excessive noise, while dynamic $\delta(t)$ is better.

\subsection{Details of the flow models} \label{app:b}
\subsubsection{Flow Matching (FM)}

Flow Matching (FM) is based on continuous normalizing flows, where the generative model is defined as an ordinary differential equation (ODE):

\begin{equation}
    \frac{dz_t}{dt} = v(z_t, t)
\end{equation}

Here, $t \in [0, 1]$, and $v(z_t, t)$ is referred to as the vector field. This ODE defines a probability path $p_t$, transitioning from a noise distribution $p_1$ to the data distribution $p_0$. Once $v(z_t, t)$ is known, we can use an ODE solver to generate data samples.

The vector field $v_\theta(z_t, t)$ is parameterized by a neural network, and the FM objective is:
\begin{equation}
L_{FM} = \mathbb{E}_{t, p_t(z)} \| v_\theta(z, t) - u_t(z) \|^2_2
\end{equation}

Here, $u_t(z)$ is the target vector field, defining the probability path $p_t(z)$ from $p_1$ to $p_0$. Without priors, $u_t(z)$ is unknown. FM introduces a predefined $u_t(z)$ using a conditional probability path $p_t(z|x_0)$, where $x_0$ is the real data. This is modeled as:
\begin{equation}
p_t(z|x_0) = \mathcal{N}(z | a_t x_0, b_t^2 I)
\end{equation}
Here, $a_t$ and $b_t$ are time-dependent functions. If $a_0 = 1, b_0 = 0$, then $p_0(z|x_0) = q(x_0)$, and if $a_1 = 0, b_1 = 1$, then $p_1(z|x_0) = p_1$. This aligns with the diffusion process $z_t = a_t x_0 + b_t \epsilon$, where $\epsilon \sim \mathcal{N}(0, I)$.

\subsubsection{Conditional Flow Matching (CFM)}

The Conditional Flow Matching (CFM) objective is:
\begin{equation}
L_{CFM} = \mathbb{E}_{t, q(x_0), p_t(z|x_0)} \| v_\theta(z, t) - u_t(z|x_0) \|^2_2
\end{equation}

Here, $u_t(z|x_0)$ defines the conditional probability path $p_t(z|x_0)$. It can be derived as:

\begin{equation}
u_t(z|x_0) = \frac{a_t'}{a_t} z_t - b_t^2 \lambda_t' \epsilon
\end{equation}

where $\lambda_t = \log \frac{a_t^2}{b_t^2}$ is the Signal-to-Noise Ratio(SNR), then we ca get $\lambda_t' = 2 \left(\frac{a_t'}{a_t} - \frac{b_t'}{b_t}\right)$. The final CFM objective becomes:

{\small
\begin{equation}
L_{CFM} = \mathbb{E}_{t, q(x_0), p_t(z|x_0), \epsilon \sim \mathcal{N}(0, I)} \left\| v_\theta(z, t) - \left( \frac{a_t'}{a_t} z_t - b_t^2 \lambda_t' \epsilon \right) \right\|^2_2
\end{equation}
}

By defining $v_\theta(z, t) = \frac{a_t'}{a_t} z_t - b_t^2 \lambda_t' \epsilon_\theta(z, t)$, this becomes equivalent to noise prediction like DDPM, with an additional weight term.

\subsubsection{Rectified Flow (RF)}

RF simplifies the forward process to:

\[
z_t = (1 - t) x_0 + t \epsilon
\]

This linear interpolation allows larger step sizes during sampling. Its loss is straightforward:

\[
L_{RF} = \mathbb{E}_{t, q(x_0), p_t(z|x_0), \epsilon \sim \mathcal{N}(0, I)} \| v_\theta(z, t) - (\epsilon - x_0) \|^2_2
\]

For RF, \( w_t = \frac{t}{1 - t} \), showing its connection to FM and other diffusion models.

\subsection{Implementation details}
All experiments are conducted on NVIDIA GeForce RTX V100 GPU and NVIDIA RTX A100 GPU for image and video sampling. We obtain all the necessary weights from publicly available repositories.

\subsubsection{Metrics for Text-to-Image Generation}
To evaluate the performance of Text-to-Image models, we employed several metrics: 
\begin{itemize}
    \item \textbf{Fréchet Inception Distance (FID)} \cite{heusel2018ganstrainedtimescaleupdate} quantifies the similarity between generated images and real images.
    \item \textbf{Human Preference Score (HPS v2.1)} \cite{wu2023humanpreferencescorev2} assesses the quality of generated images based on human judgment.
    \item \textbf{CLIP Score} \cite{taited2023CLIPScore} measures the alignment between generated images and their textual descriptions.
    \item \textbf{Aesthetic Score} \cite{improvedaestheticpredictor} evaluates the visual quality and appeal of the images.
\end{itemize}
FID, CLIP Score, and Aesthetic Score were conducted using 5000 prompts in the MS-COCO 2017 validation dataset. HPS v2.1 used the standard prompts they provided.
\begin{table*}
    \centering
    \caption{The parameters of CFG, PAG, and SG used during inference for four diffusion models.}
    \begin{tabular}{ccccc}
    \toprule
          &  Stable-Diffusion 1.4&  Stable-Diffusion 3-Medium&  Stable-Diffusion 3.5-Large& Flux.1-dev\\
          \midrule
         CFG guidance scale&  7.5&  3.5&  3.5& 3.5\\
         PAG guidance scale&  0.3&  0.7&  -& -\\
         SG guidance scale&  3&  3&  3& 3\\
 SG constant shift scale& 10& 10& 10&10\\
 \bottomrule
    \end{tabular}
    
    \label{tab:parameters}
\end{table*}
\subsubsection{Metrics for Text-to-Video Generation}
To evaluate the performance of Text-to-Video models, we utilized VBench\cite{huang2023vbenchcomprehensivebenchmarksuite}, a comprehensive benchmark specifically designed for video generation. We evaluate model performance across five key dimensions: human action, scene, multiple objects, appearance style, and overall consistency. 
\begin{itemize}
    \item \textbf{Human Action:} Use UMT \cite{li2023unmasked} to assess if generated videos accurately depict the specific actions described in text prompts.
    \item \textbf{Scene:} Use Tag2Text \cite{huang2024tag2textguidingvisionlanguagemod} to caption generated scenes and verify their consistency with the text prompt (e.g., "ocean" vs. "river").
    \item \textbf{Multiple Objects:} Evaluate the success rate of generating all objects specified in the text prompt within each video frame.
    \item \textbf{Appearance Style:} Use CLIP to feature similarity between frames and style descriptions (e.g., "oil painting" or "cyberpunk").
    \item \textbf{Overall Consistency:} Use ViCLIP \cite{wang2024internvidlargescalevideotextdataset} to calculate overall video-text consistency, reflecting semantic and style alignment.
\end{itemize}

\subsubsection{Guidance Scale Setting} \label{app:c3}
The guidance scales for CFG and PAG are kept the same as the original settings in the corresponding papers. For each diffusion model, we provide the best guidance scale and constant shift scale of Self-Guidance. The specific parameters for these three guidance methods are in Table \ref{tab:parameters}.

\subsection{Limitations}
While Self-Guidance (SG) and its efficient variant SG-prev offer strong improvements, there are several limitations to consider. First, SG requires two forward passes per sampling step, which increases inference time. Although SG-prev mitigates this overhead, its approximation may not capture all features in complex scenes, leading to slightly reduced performance in several cases. 

Second, SG is limited in its ability to make global corrections. If a generated image contains high-level structural errors, SG is often insufficient to induce large-scale modifications. As shown in Fig.~\ref{fig:limitations}, these examples include cases where generated images have high-level structural errors such as:  
1) Large-scale missing body parts (e.g., entire arms or legs missing),  
2) Multiple objects overlapping unnaturally,  
3) Text generation errors (e.g., incorrect or illegible text in the scene), and  
4) Objects with distorted perspective (e.g., incorrect spatial relationships or warped shapes).

Third, as our work primarily focuses on the Text-to-Image task, the Text-to-Video experiments are relatively preliminary. We will pay more attention to the Text-to-Video setting for our future work.

%% file: main.bbl
\begin{thebibliography}{69}
\providecommand{\natexlab}[1]{#1}
\providecommand{\url}[1]{\texttt{#1}}
\expandafter\ifx\csname urlstyle\endcsname\relax
  \providecommand{\doi}[1]{doi: #1}\else
  \providecommand{\doi}{doi: \begingroup \urlstyle{rm}\Url}\fi

\bibitem[Ahn et~al.(2024)Ahn, Cho, Min, Jang, Kim, Kim, Park, Jin, and Kim]{ahn2024selfrectifyingdiffusionsamplingperturbedattention}
Donghoon Ahn, Hyoungwon Cho, Jaewon Min, Wooseok Jang, Jungwoo Kim, SeonHwa Kim, Hyun~Hee Park, Kyong~Hwan Jin, and Seungryong Kim.
\newblock Self-rectifying diffusion sampling with perturbed-attention guidance, 2024.

\bibitem[AI(2023)]{stability2023stable_diffusion}
Stability AI.
\newblock Introducing stable diffusion 3.5, 2023.
\newblock Accessed: 2024-10-30.

\bibitem[Alemohammad et~al.(2024)Alemohammad, Humayun, Agarwal, Collomosse, and Baraniuk]{alemohammad2024selfimprovingdiffusionmodelssynthetic}
Sina Alemohammad, Ahmed~Imtiaz Humayun, Shruti Agarwal, John Collomosse, and Richard Baraniuk.
\newblock Self-improving diffusion models with synthetic data, 2024.

\bibitem[Blattmann et~al.(2023)Blattmann, Dockhorn, Kulal, Mendelevitch, Kilian, Lorenz, Levi, English, Voleti, Letts, Jampani, and Rombach]{blattmann2023stablevideodiffusionscaling}
Andreas Blattmann, Tim Dockhorn, Sumith Kulal, Daniel Mendelevitch, Maciej Kilian, Dominik Lorenz, Yam Levi, Zion English, Vikram Voleti, Adam Letts, Varun Jampani, and Robin Rombach.
\newblock Stable video diffusion: Scaling latent video diffusion models to large datasets, 2023.

\bibitem[Chen et~al.(2019)Chen, Behrmann, Duvenaud, and Jacobsen]{chen2019residual}
Ricky~TQ Chen, Jens Behrmann, David~K Duvenaud, and J{\"o}rn-Henrik Jacobsen.
\newblock Residual flows for invertible generative modeling.
\newblock In \emph{Advances in Neural Information Processing Systems}, pages 9916--9926, 2019.

\bibitem[Couairon et~al.(2022)Couairon, Verbeek, Schwenk, and Cord]{Couairon2022DiffEditDS}
Guillaume Couairon, Jakob Verbeek, Holger Schwenk, and Matthieu Cord.
\newblock Diffedit: Diffusion-based semantic image editing with mask guidance.
\newblock \emph{ArXiv}, abs/2210.11427, 2022.

\bibitem[Deng et~al.(2024)Deng, Luo, Tan, Bilo{\v{s}}, Chen, Nevmyvaka, and Chen]{deng2024variational}
Wei Deng, Weijian Luo, Yixin Tan, Marin Bilo{\v{s}}, Yu Chen, Yuriy Nevmyvaka, and Ricky~TQ Chen.
\newblock Variational schr$\backslash$" odinger diffusion models.
\newblock \emph{arXiv preprint arXiv:2405.04795}, 2024.

\bibitem[Esser et~al.(2024)Esser, Kulal, Blattmann, Entezari, Müller, Saini, Levi, Lorenz, Sauer, Boesel, Podell, Dockhorn, English, Lacey, Goodwin, Marek, and Rombach]{esser2024scalingrectifiedflowtransformers}
Patrick Esser, Sumith Kulal, Andreas Blattmann, Rahim Entezari, Jonas Müller, Harry Saini, Yam Levi, Dominik Lorenz, Axel Sauer, Frederic Boesel, Dustin Podell, Tim Dockhorn, Zion English, Kyle Lacey, Alex Goodwin, Yannik Marek, and Robin Rombach.
\newblock Scaling rectified flow transformers for high-resolution image synthesis, 2024.

\bibitem[Geng et~al.(2024)Geng, Pokle, Luo, Lin, and Kolter]{geng2024consistency}
Zhengyang Geng, Ashwini Pokle, William Luo, Justin Lin, and J~Zico Kolter.
\newblock Consistency models made easy.
\newblock \emph{arXiv preprint arXiv:2406.14548}, 2024.

\bibitem[Heusel et~al.(2018)Heusel, Ramsauer, Unterthiner, Nessler, and Hochreiter]{heusel2018ganstrainedtimescaleupdate}
Martin Heusel, Hubert Ramsauer, Thomas Unterthiner, Bernhard Nessler, and Sepp Hochreiter.
\newblock Gans trained by a two time-scale update rule converge to a local nash equilibrium, 2018.

\bibitem[Ho and Salimans(2022)]{ho2022classifierfreediffusionguidance}
Jonathan Ho and Tim Salimans.
\newblock Classifier-free diffusion guidance, 2022.

\bibitem[Ho et~al.(2020{\natexlab{a}})Ho, Jain, and Abbeel]{NEURIPS2020_4c5bcfec}
Jonathan Ho, Ajay Jain, and Pieter Abbeel.
\newblock Denoising diffusion probabilistic models.
\newblock \emph{Advances in Neural Information Processing Systems}, 33:\penalty0 6840--6851, 2020{\natexlab{a}}.

\bibitem[Ho et~al.(2020{\natexlab{b}})Ho, Jain, and Abbeel]{ho2020denoising}
Jonathan Ho, Ajay Jain, and Pieter Abbeel.
\newblock Denoising diffusion probabilistic models.
\newblock \emph{Advances in Neural Information Processing Systems}, 33:\penalty0 6840--6851, 2020{\natexlab{b}}.

\bibitem[Ho et~al.(2022)Ho, Salimans, Gritsenko, Chan, Norouzi, and Fleet]{ho2022video}
Jonathan Ho, Tim Salimans, Alexey Gritsenko, William Chan, Mohammad Norouzi, and David~J Fleet.
\newblock Video diffusion models.
\newblock \emph{arXiv preprint arXiv:2204.03458}, 2022.

\bibitem[Hong(2024)]{hong2024smoothedenergyguidanceguiding}
Susung Hong.
\newblock Smoothed energy guidance: Guiding diffusion models with reduced energy curvature of attention, 2024.

\bibitem[Hong et~al.(2023)Hong, Lee, Jang, and Kim]{hong2023improvingsamplequalitydiffusion}
Susung Hong, Gyuseong Lee, Wooseok Jang, and Seungryong Kim.
\newblock Improving sample quality of diffusion models using self-attention guidance, 2023.

\bibitem[Hoogeboom et~al.(2022)Hoogeboom, Satorras, Vignac, and Welling]{hoogeboom2022equivariant}
Emiel Hoogeboom, V{\i}ctor~Garcia Satorras, Cl{\'e}ment Vignac, and Max Welling.
\newblock Equivariant diffusion for molecule generation in 3d.
\newblock In \emph{International Conference on Machine Learning}, pages 8867--8887. PMLR, 2022.

\bibitem[Huang et~al.(2024{\natexlab{a}})Huang, Zhang, Ma, Tian, Feng, Zhang, Li, Guo, and Zhang]{huang2024tag2textguidingvisionlanguagemod}
Xinyu Huang, Youcai Zhang, Jinyu Ma, Weiwei Tian, Rui Feng, Yuejie Zhang, Yaqian Li, Yandong Guo, and Lei Zhang.
\newblock Tag2text: Guiding vision-language model via image tagging, 2024{\natexlab{a}}.

\bibitem[Huang et~al.(2023)Huang, He, Yu, Zhang, Si, Jiang, Zhang, Wu, Jin, Chanpaisit, Wang, Chen, Wang, Lin, Qiao, and Liu]{huang2023vbenchcomprehensivebenchmarksuite}
Ziqi Huang, Yinan He, Jiashuo Yu, Fan Zhang, Chenyang Si, Yuming Jiang, Yuanhan Zhang, Tianxing Wu, Qingyang Jin, Nattapol Chanpaisit, Yaohui Wang, Xinyuan Chen, Limin Wang, Dahua Lin, Yu Qiao, and Ziwei Liu.
\newblock Vbench: Comprehensive benchmark suite for video generative models, 2023.

\bibitem[Huang et~al.(2024{\natexlab{b}})Huang, Geng, Luo, and Qi]{huang2024flow}
Zemin Huang, Zhengyang Geng, Weijian Luo, and Guo-jun Qi.
\newblock Flow generator matching.
\newblock \emph{arXiv preprint arXiv:2410.19310}, 2024{\natexlab{b}}.

\bibitem[Karras et~al.(2020)Karras, Laine, Aittala, Hellsten, Lehtinen, and Aila]{karras2020analyzing}
Tero Karras, Samuli Laine, Miika Aittala, Janne Hellsten, Jaakko Lehtinen, and Timo Aila.
\newblock Analyzing and improving the image quality of stylegan.
\newblock In \emph{Proceedings of the IEEE/CVF conference on computer vision and pattern recognition}, pages 8110--8119, 2020.

\bibitem[Karras et~al.(2022)Karras, Aittala, Aila, and Laine]{karras2022edm}
Tero Karras, Miika Aittala, Timo Aila, and Samuli Laine.
\newblock Elucidating the design space of diffusion-based generative models.
\newblock In \emph{Proc. NeurIPS}, 2022.

\bibitem[Karras et~al.(2024{\natexlab{a}})Karras, Aittala, Kynkäänniemi, Lehtinen, Aila, and Laine]{karras2024guidingdiffusionmodelbad}
Tero Karras, Miika Aittala, Tuomas Kynkäänniemi, Jaakko Lehtinen, Timo Aila, and Samuli Laine.
\newblock Guiding a diffusion model with a bad version of itself, 2024{\natexlab{a}}.

\bibitem[Karras et~al.(2024{\natexlab{b}})Karras, Aittala, Lehtinen, Hellsten, Aila, and Laine]{karras2024analyzingimprovingtrainingdynamics}
Tero Karras, Miika Aittala, Jaakko Lehtinen, Janne Hellsten, Timo Aila, and Samuli Laine.
\newblock Analyzing and improving the training dynamics of diffusion models, 2024{\natexlab{b}}.

\bibitem[Kim et~al.(2022)Kim, Kim, and Yoon]{kim2021guidedtts}
Heeseung Kim, Sungwon Kim, and Sungroh Yoon.
\newblock Guided-tts: A diffusion model for text-to-speech via classifier guidance.
\newblock In \emph{International Conference on Machine Learning}, pages 11119--11133. PMLR, 2022.

\bibitem[Kingma and Dhariwal(2018)]{kingma2018glowgenerativeflowinvertible}
Diederik~P. Kingma and Prafulla Dhariwal.
\newblock Glow: Generative flow with invertible 1x1 convolutions, 2018.

\bibitem[Kirstain et~al.(2023)Kirstain, Polyak, Singer, Matiana, Penna, and Levy]{kirstain2023pickapicopendatasetuser}
Yuval Kirstain, Adam Polyak, Uriel Singer, Shahbuland Matiana, Joe Penna, and Omer Levy.
\newblock Pick-a-pic: An open dataset of user preferences for text-to-image generation, 2023.

\bibitem[Kong et~al.(2020)Kong, Ping, Huang, Zhao, and Catanzaro]{kong2021diffwave}
Zhifeng Kong, Wei Ping, Jiaji Huang, Kexin Zhao, and Bryan Catanzaro.
\newblock Diffwave: A versatile diffusion model for audio synthesis.
\newblock In \emph{International Conference on Learning Representations}, 2020.

\bibitem[Labs(2023)]{blackforestlabs}
Black~Forest Labs.
\newblock Flux.1, 2023.

\bibitem[Li et~al.(2023)Li, Wang, Li, Wang, He, Wang, and Qiao]{li2023unmasked}
Kunchang Li, Yali Wang, Yizhuo Li, Yi Wang, Yinan He, Limin Wang, and Yu Qiao.
\newblock Unmasked teacher: Towards training-efficient video foundation models.
\newblock In \emph{Proceedings of the IEEE/CVF International Conference on Computer Vision}, pages 19948--19960, 2023.

\bibitem[Liao et~al.(2024)Liao, Xie, Chen, Lu, and Deng]{liao2024facescorebenchmarkingenhancingface}
Zhenyi Liao, Qingsong Xie, Chen Chen, Hannan Lu, and Zhijie Deng.
\newblock Facescore: Benchmarking and enhancing face quality in human generation, 2024.

\bibitem[Lipman et~al.(2022)Lipman, Chen, Ben-Hamu, Nickel, and Le]{Lipman2022FlowMF}
Yaron Lipman, Ricky T.~Q. Chen, Heli Ben-Hamu, Maximilian Nickel, and Matt Le.
\newblock Flow matching for generative modeling.
\newblock \emph{ArXiv}, abs/2210.02747, 2022.

\bibitem[Liu et~al.(2022{\natexlab{a}})Liu, Ren, Lin, and Zhao]{liu2022pseudo}
Luping Liu, Yi Ren, Zhijie Lin, and Zhou Zhao.
\newblock Pseudo numerical methods for diffusion models on manifolds.
\newblock \emph{arXiv preprint arXiv:2202.09778}, 2022{\natexlab{a}}.

\bibitem[Liu et~al.(2022{\natexlab{b}})Liu, Gong, and Liu]{liu2022flow}
Xingchao Liu, Chengyue Gong, and Qiang Liu.
\newblock Flow straight and fast: Learning to generate and transfer data with rectified flow.
\newblock \emph{arXiv preprint arXiv:2209.03003}, 2022{\natexlab{b}}.

\bibitem[Lu et~al.(2022)Lu, Zhou, Bao, Chen, Li, and Zhu]{lu2022dpm}
Cheng Lu, Yuhao Zhou, Fan Bao, Jianfei Chen, Chongxuan Li, and Jun Zhu.
\newblock Dpm-solver: A fast ode solver for diffusion probabilistic model sampling in around 10 steps.
\newblock \emph{arXiv preprint arXiv:2206.00927}, 2022.

\bibitem[Luo(2024)]{luo2024diffpp}
Weijian Luo.
\newblock Diff-instruct++: Training one-step text-to-image generator model to align with human preferences.
\newblock \emph{arXiv preprint arXiv:2410.18881}, 2024.

\bibitem[Luo et~al.(2024{\natexlab{a}})Luo, Hu, Zhang, Sun, Li, and Zhang]{luo2024diffinstruct}
Weijian Luo, Tianyang Hu, Shifeng Zhang, Jiacheng Sun, Zhenguo Li, and Zhihua Zhang.
\newblock Diff-instruct: A universal approach for transferring knowledge from pre-trained diffusion models.
\newblock \emph{Advances in Neural Information Processing Systems}, 36, 2024{\natexlab{a}}.

\bibitem[Luo et~al.(2024{\natexlab{b}})Luo, Huang, Geng, Kolter, and Qi]{luo2024one}
Weijian Luo, Zemin Huang, Zhengyang Geng, J~Zico Kolter, and Guo-jun Qi.
\newblock One-step diffusion distillation through score implicit matching.
\newblock \emph{arXiv preprint arXiv:2410.16794}, 2024{\natexlab{b}}.

\bibitem[Luo et~al.(2024{\natexlab{c}})Luo, Zhang, and Zhang]{luo2024entropy}
Weijian Luo, Boya Zhang, and Zhihua Zhang.
\newblock Entropy-based training methods for scalable neural implicit samplers.
\newblock \emph{Advances in Neural Information Processing Systems}, 36, 2024{\natexlab{c}}.

\bibitem[Luo et~al.(2024{\natexlab{d}})Luo, Zhang, Zhang, and Geng]{luo2024diffstar}
Weijian Luo, Colin Zhang, Debing Zhang, and Zhengyang Geng.
\newblock Diff-instruct*: Towards human-preferred one-step text-to-image generative models.
\newblock \emph{arXiv preprint arXiv:2410.20898}, 2024{\natexlab{d}}.

\bibitem[Luo et~al.(2024{\natexlab{e}})Luo, Zhang, Qiu, Yao, Chen, Jiang, and Mei]{10.1145/3664647.3681506}
Yang Luo, Yiheng Zhang, Zhaofan Qiu, Ting Yao, Zhineng Chen, Yu-Gang Jiang, and Tao Mei.
\newblock Freeenhance: Tuning-free image enhancement via content-consistent noising-and-denoising process.
\newblock In \emph{Proceedings of the 32nd ACM International Conference on Multimedia}, page 7075–7084, New York, NY, USA, 2024{\natexlab{e}}. Association for Computing Machinery.

\bibitem[Meng et~al.(2021)Meng, Song, Song, Wu, Zhu, and Ermon]{meng2021sdedit}
Chenlin Meng, Yang Song, Jiaming Song, Jiajun Wu, Jun-Yan Zhu, and Stefano Ermon.
\newblock Sdedit: Image synthesis and editing with stochastic differential equations.
\newblock \emph{arXiv preprint arXiv:2108.01073}, 2021.

\bibitem[Nagolinc(2023)]{NagolincHandCaptions}
Nagolinc.
\newblock Hand captions, 2023.
\newblock Accessed: 2024-10-31.

\bibitem[Nichol and Dhariwal(2021)]{nichol2021improved}
Alex Nichol and Prafulla Dhariwal.
\newblock Improved denoising diffusion probabilistic models.
\newblock \emph{arXiv preprint arXiv:2102.09672}, 2021.

\bibitem[Oord et~al.(2016)Oord, Dieleman, Zen, Simonyan, Vinyals, Graves, Kalchbrenner, Senior, and Kavukcuoglu]{oord2016wavenet}
Aaron van~den Oord, Sander Dieleman, Heiga Zen, Karen Simonyan, Oriol Vinyals, Alex Graves, Nal Kalchbrenner, Andrew Senior, and Koray Kavukcuoglu.
\newblock Wavenet: A generative model for raw audio.
\newblock \emph{arXiv preprint arXiv:1609.03499}, 2016.

\bibitem[OpenFace and CQUPT(2023)]{OpenFaceCQUPTHumanCaption10M}
OpenFace and CQUPT.
\newblock Humancaption-10m, 2023.
\newblock Accessed: 2024-10-31.

\bibitem[Parmar et~al.(2022)Parmar, Zhang, and Zhu]{parmar2022aliasedresizingsurprisingsubtleties}
Gaurav Parmar, Richard Zhang, and Jun-Yan Zhu.
\newblock On aliased resizing and surprising subtleties in gan evaluation, 2022.

\bibitem[Poole et~al.(2022)Poole, Jain, Barron, and Mildenhall]{poole2022dreamfusion}
Ben Poole, Ajay Jain, Jonathan~T Barron, and Ben Mildenhall.
\newblock Dreamfusion: Text-to-3d using 2d diffusion.
\newblock \emph{arXiv preprint arXiv:2209.14988}, 2022.

\bibitem[Ramesh et~al.(2021)Ramesh, Pavlov, Goh, Gray, Voss, Radford, Chen, and Sutskever]{ramesh2021zero}
Aditya Ramesh, Mikhail Pavlov, Gabriel Goh, Scott Gray, Chelsea Voss, Alec Radford, Mark Chen, and Ilya Sutskever.
\newblock Zero-shot text-to-image generation.
\newblock In \emph{International Conference on Machine Learning}, pages 8821--8831. PMLR, 2021.

\bibitem[Ramesh et~al.(2022)Ramesh, Dhariwal, Nichol, Chu, and Chen]{ramesh2022hierarchical}
Aditya Ramesh, Prafulla Dhariwal, Alex Nichol, Casey Chu, and Mark Chen.
\newblock Hierarchical text-conditional image generation with clip latents.
\newblock \emph{arXiv preprint arXiv:2204.06125}, 2022.

\bibitem[Rombach et~al.(2022)Rombach, Blattmann, Lorenz, Esser, and Ommer]{rombach2022high}
Robin Rombach, Andreas Blattmann, Dominik Lorenz, Patrick Esser, and Bj{\"o}rn Ommer.
\newblock High-resolution image synthesis with latent diffusion models.
\newblock In \emph{Proceedings of the IEEE/CVF Conference on Computer Vision and Pattern Recognition}, pages 10684--10695, 2022.

\bibitem[Sadat et~al.(2024)Sadat, Kansy, Hilliges, and Weber]{sadat2024trainingproblemrethinkingclassifierfree}
Seyedmorteza Sadat, Manuel Kansy, Otmar Hilliges, and Romann~M. Weber.
\newblock No training, no problem: Rethinking classifier-free guidance for diffusion models, 2024.

\bibitem[Saharia et~al.(2022)Saharia, Chan, Saxena, Li, Whang, Denton, Ghasemipour, Ayan, Mahdavi, Lopes, et~al.]{saharia2022photorealistic}
Chitwan Saharia, William Chan, Saurabh Saxena, Lala Li, Jay Whang, Emily Denton, Seyed Kamyar~Seyed Ghasemipour, Burcu~Karagol Ayan, S~Sara Mahdavi, Rapha~Gontijo Lopes, et~al.
\newblock Photorealistic text-to-image diffusion models with deep language understanding.
\newblock \emph{arXiv preprint arXiv:2205.11487}, 2022.

\bibitem[Schuhmann(2023)]{improvedaestheticpredictor}
Christoph Schuhmann.
\newblock Improved aesthetic predictor, 2023.

\bibitem[Shen et~al.(2023)Shen, Zhao, Meng, Li, Zhu, Zhou, and Lu]{shen2023difftalkcraftingdiffusionmodels}
Shuai Shen, Wenliang Zhao, Zibin Meng, Wanhua Li, Zheng Zhu, Jie Zhou, and Jiwen Lu.
\newblock Difftalk: Crafting diffusion models for generalized audio-driven portraits animation, 2023.

\bibitem[Sohl-Dickstein et~al.(2009)Sohl-Dickstein, Battaglino, and DeWeese]{Dickstein11}
Jascha Sohl-Dickstein, Peter Battaglino, and Michael~R DeWeese.
\newblock Minimum probability flow learning.
\newblock \emph{arXiv preprint arXiv:0906.4779}, 2009.

\bibitem[Song et~al.(2020{\natexlab{a}})Song, Meng, and Ermon]{song2020denoising}
Jiaming Song, Chenlin Meng, and Stefano Ermon.
\newblock Denoising diffusion implicit models.
\newblock \emph{arXiv preprint arXiv:2010.02502}, 2020{\natexlab{a}}.

\bibitem[Song et~al.(2020{\natexlab{b}})Song, Sohl-Dickstein, Kingma, Kumar, Ermon, and Poole]{song2021scorebased}
Yang Song, Jascha Sohl-Dickstein, Diederik~P Kingma, Abhishek Kumar, Stefano Ermon, and Ben Poole.
\newblock Score-based generative modeling through stochastic differential equations.
\newblock In \emph{International Conference on Learning Representations}, 2020{\natexlab{b}}.

\bibitem[Song et~al.(2021)Song, Durkan, Murray, and Ermon]{NEURIPS2021_0a9fdbb1}
Yang Song, Conor Durkan, Iain Murray, and Stefano Ermon.
\newblock Maximum likelihood training of score-based diffusion models.
\newblock \emph{Advances in Neural Information Processing Systems}, 34:\penalty0 1415--1428, 2021.

\bibitem[Vincent(2011)]{vincent2011connection}
Pascal Vincent.
\newblock A {C}onnection {B}etween {S}core {M}atching and {D}enoising {A}utoencoders.
\newblock \emph{Neural Computation}, 23\penalty0 (7):\penalty0 1661--1674, 2011.

\bibitem[Wang et~al.(2024{\natexlab{a}})Wang, Bai, Luo, Chen, and Sun]{wang2024integrating}
Yifei Wang, Weimin Bai, Weijian Luo, Wenzheng Chen, and He Sun.
\newblock Integrating amortized inference with diffusion models for learning clean distribution from corrupted images.
\newblock \emph{arXiv preprint arXiv:2407.11162}, 2024{\natexlab{a}}.

\bibitem[Wang et~al.(2024{\natexlab{b}})Wang, He, Li, Li, Yu, Ma, Li, Chen, Chen, Wang, He, Luo, Liu, Wang, Wang, and Qiao]{wang2024internvidlargescalevideotextdataset}
Yi Wang, Yinan He, Yizhuo Li, Kunchang Li, Jiashuo Yu, Xin Ma, Xinhao Li, Guo Chen, Xinyuan Chen, Yaohui Wang, Conghui He, Ping Luo, Ziwei Liu, Yali Wang, Limin Wang, and Yu Qiao.
\newblock Internvid: A large-scale video-text dataset for multimodal understanding and generation, 2024{\natexlab{b}}.

\bibitem[Wu et~al.(2023)Wu, Hao, Sun, Chen, Zhu, Zhao, and Li]{wu2023humanpreferencescorev2}
Xiaoshi Wu, Yiming Hao, Keqiang Sun, Yixiong Chen, Feng Zhu, Rui Zhao, and Hongsheng Li.
\newblock Human preference score v2: A solid benchmark for evaluating human preferences of text-to-image synthesis, 2023.

\bibitem[Xu et~al.(2023)Xu, Liu, Wu, Tong, Li, Ding, Tang, and Dong]{xu2023imagerewardlearningevaluatinghuman}
Jiazheng Xu, Xiao Liu, Yuchen Wu, Yuxuan Tong, Qinkai Li, Ming Ding, Jie Tang, and Yuxiao Dong.
\newblock Imagereward: Learning and evaluating human preferences for text-to-image generation, 2023.

\bibitem[Xue et~al.(2024)Xue, Yi, Luo, Zhang, Sun, Li, and Ma]{xue2024sa}
Shuchen Xue, Mingyang Yi, Weijian Luo, Shifeng Zhang, Jiacheng Sun, Zhenguo Li, and Zhi-Ming Ma.
\newblock Sa-solver: Stochastic adams solver for fast sampling of diffusion models.
\newblock \emph{Advances in Neural Information Processing Systems}, 36, 2024.

\bibitem[Yang et~al.(2024)Yang, Teng, Zheng, Ding, Huang, Xu, Yang, Hong, Zhang, Feng, Yin, Gu, Zhang, Wang, Cheng, Liu, Xu, Dong, and Tang]{yang2024cogvideoxtexttovideodiffusionmodels}
Zhuoyi Yang, Jiayan Teng, Wendi Zheng, Ming Ding, Shiyu Huang, Jiazheng Xu, Yuanming Yang, Wenyi Hong, Xiaohan Zhang, Guanyu Feng, Da Yin, Xiaotao Gu, Yuxuan Zhang, Weihan Wang, Yean Cheng, Ting Liu, Bin Xu, Yuxiao Dong, and Jie Tang.
\newblock Cogvideox: Text-to-video diffusion models with an expert transformer, 2024.

\bibitem[Zhang et~al.(2023)Zhang, Luo, and Zhang]{zhang2023enhancing}
Boya Zhang, Weijian Luo, and Zhihua Zhang.
\newblock Enhancing adversarial robustness via score-based optimization.
\newblock \emph{Advances in Neural Information Processing Systems}, 36:\penalty0 51810--51829, 2023.

\bibitem[Zhang et~al.(2020)Zhang, Bazarevsky, Vakunov, Tkachenka, Sung, Chang, and Grundmann]{zhang2020mediapipehandsondevicerealtime}
Fan Zhang, Valentin Bazarevsky, Andrey Vakunov, Andrei Tkachenka, George Sung, Chuo-Ling Chang, and Matthias Grundmann.
\newblock Mediapipe hands: On-device real-time hand tracking, 2020.

\bibitem[Zhengwentai(2023)]{taited2023CLIPScore}
SUN Zhengwentai.
\newblock {clip-score: CLIP Score for PyTorch}.
\newblock \url{https://github.com/taited/clip-score}, 2023.
\newblock Version 0.1.1.

\end{thebibliography}
